\documentclass[runningheads]{llncs}
\usepackage{graphicx}
\usepackage{amsmath,amssymb} 
\usepackage{color}
\usepackage[width=122mm,left=12mm,paperwidth=146mm,height=193mm,top=12mm,paperheight=217mm]{geometry}
\usepackage{epsfig}
\usepackage{epstopdf}
\usepackage[dvipsnames]{xcolor}
\usepackage{amsmath,amsfonts}
\usepackage{amssymb,amsopn}
\usepackage{bm} 

\newcommand{\vct}[1]{\boldsymbol{#1}} 
\newcommand{\cst}[1]{\mathsf{#1}}  

\newcommand{\field}[1]{\mathbb{#1}}
\newcommand{\R}{\field{R}} 
\newcommand{\I}{\field{I}} 
\newcommand{\T}{^{\textrm T}} 

\newcommand{\ProbOpr}[1]{\mathbb{#1}}

\newcommand{\expect}[2]{\ProbOpr{E}_{#1}\left[#2\right]}
\newcommand{\var}[2]{%
\ifthenelse{\equal{#2}{}}{\ProbOpr{VAR}_{#1}}
{\ifthenelse{\equal{#1}{}}{\ProbOpr{VAR}\left[#2\right]}{\ProbOpr{VAR}_{#1}\left[#2\right]}}} 

\newcommand{\vx}{{\vct{x}}}

\newcommand{\va}{\vct{a}}

\newcommand{\vw}{\vct{w}}

\newcommand{\cN}{\cst{N}}

\newcommand{\cD}{\cst{D}}

\newcommand{\cU}{\cst{U}}
\newcommand{\cS}{\cst{S}}

\newcommand{\sS}{\mathcal{S}}

\newcommand{\eat}[1]{}

\usepackage[pagebackref=true,breaklinks=true,colorlinks,bookmarks=false]{hyperref}

\begin{document}
\pagestyle{headings}
\mainmatter

\title{An Empirical Study and Analysis of \\Generalized Zero-Shot Learning \\for Object Recognition in the Wild}

\titlerunning{An Empirical Study \& Analysis of GZSL for Object Recognition in the Wild}
\authorrunning{Wei-Lun Chao$^*$, Soravit Changpinyo$^*$, Boqing Gong, and Fei Sha}

\author{Wei-Lun Chao$^*$$^1$, Soravit Changpinyo\thanks{\hspace{4pt}Equal contribution.}$^1$, Boqing Gong$^2$, and Fei Sha$^3$}
\institute{$^1$Dept. of Computer Science, U. of Southern California, United States\\
$^2$Center for Research in Computer Vision, U. of Central Florida, United States\\
$^3$Dept. of Computer Science, U. of California, Los Angeles, United States\\
\email{\{weilunc, schangpi\}@usc.edu, bgong@crcv.ucf.edu, feisha@cs.ucla.edu}}

\maketitle


\begin{abstract}

\emph{Zero-shot learning} (ZSL) methods have been studied in the unrealistic setting where test data are assumed to come from unseen classes only. In this paper, we advocate studying the problem of \emph{generalized zero-shot learning} (GZSL) where the test data's class memberships are unconstrained. We show empirically that naively using the classifiers constructed by ZSL approaches does not perform well in the generalized setting. Motivated by this, we propose a simple but effective calibration method that can be used to balance two conflicting forces: recognizing data from seen classes versus those from unseen ones. We develop a performance metric to characterize such a trade-off and examine the utility of this metric in evaluating various ZSL approaches. Our analysis further shows that there is a large gap between the performance of existing approaches and an upper bound established via \emph{idealized} semantic embeddings, suggesting that improving class semantic embeddings is vital to GZSL.
\end{abstract}


\section{Introduction}
\label{sIntro}

The availability of large-scale labeled training images is one of the key factors that contribute to recent successes in visual object recognition and classification.
It is well-known, however, that object frequencies in natural images follow long-tailed distributions \cite{SudderthJ08,SalakhutdinovTT11,zhu2014capturing}.
For example, some animal or plant species are simply rare by nature --- it is uncommon to find alpacas wandering around the streets.
Furthermore, brand new categories could just emerge with zero or little labeled images; newly defined visual concepts or products are introduced everyday.       
In this \emph{real-world} setting, it would be desirable for computer vision systems to be able to recognize instances of those rare classes, while demanding minimum human efforts and labeled examples.

Zero-shot learning (ZSL) has long been believed to hold the key to the above problem of recognition in the wild. 
ZSL differentiates two types of classes: \emph{seen} and \emph{unseen}, where labeled examples are available for seen classes only. Without labeled data, models for unseen classes are learned by relating them to seen ones. 
This is often achieved by embedding both seen and unseen classes into a common semantic space, such as visual attributes \cite{LampertNH09,FarhadiEHF09,ParikhG11} or \textsc{word2vec} representations of the class names \cite{MikolovCCD13,FromeCSBDRM13,SocherGMN13}. This common semantic space enables transferring models for the seen classes to those for the unseen ones \cite{PalatucciPHM09}.

The setup for ZSL is that once models for unseen classes are learned, they are judged based on their ability to discriminate among unseen classes, assuming the absence of seen objects during the test phase.  Originally proposed in the seminal work of Lampert et al. \cite{LampertNH09}, this setting has almost always been adopted for evaluating ZSL methods \cite{PalatucciPHM09,YuA10,RohrbachSS11,KankuekulKTH12,AkataPHS13,YuCFSC13,FromeCSBDRM13,MensinkGS14,NorouziMBSSFCD14,JayaramanG14,AlHalahS15,AkataRWLS15,FuHXG15,FuXKG15,LiGS15,Bernardino15,KodirovXFG15,ZhangS15,zhang2016classifying,ChangpinyoCGS16}.   

But, \emph{does this problem setting truly reflect what recognition in the wild entails}? While the ability to learn novel concepts is by all means a trait that any zero-shot learning systems should possess, it is merely one side of the coin. The other important --- yet so far under-studied --- trait is the ability to \emph{remember} past experiences, i.e., the \emph{seen} classes. 

\emph{Why is this trait desirable}? Consider how data are distributed in the real world.
The seen classes are often more common than the unseen ones; it is therefore unrealistic to assume that we will never encounter them during the test stage. For models generated by ZSL to be truly useful, they should not only accurately discriminate among either seen {\em or} unseen classes themselves but also accurately discriminate between the seen {\em and} unseen ones.

Thus, to understand better how existing ZSL approaches will perform in the real world, we advocate evaluating them in the setting of \emph{generalized zero-shot learning} (GZSL), where test data are from both seen and unseen classes and we need to classify them into the joint labeling space of both types of classes.  Previous work in this direction is scarce. See related work for more details.

Our contributions include an extensive empirical study of several existing ZSL approaches in this new setting. We show that a straightforward application of classifiers constructed by those approaches performs poorly. In particular, test data from unseen classes are almost always classified as a class from the seen ones. We propose a surprisingly simple yet very effective method called \textit{calibrated stacking} to address this problem. This method is mindful of the two conflicting forces: recognizing data from seen classes and recognizing data from unseen ones. We introduce a new performance metric called Area Under Seen-Unseen accuracy Curve (AUSUC) that can evaluate ZSL approaches on how well they can trade off between the two. We demonstrate the utility of this metric by evaluating several representative ZSL approaches under this metric on three benchmark datasets, including the full ImageNet Fall 2011 release dataset~\cite{deng2009imagenet} that contains approximately 21,000 unseen categories. 

We complement our comparative studies in learning methods by further establishing an upper bound on the performance limit of ZSL. In particular, our idea is to use class-representative visual features as the \emph{idealized} semantic embeddings to construct ZSL classifiers. We show that there is a large gap between existing approaches and this ideal performance limit, suggesting that improving class semantic embeddings is vital to achieve GZSL.

The rest of the paper is organized as follows. Section~\ref{sRelated} reviews relevant literature. 
We define GZSL formally and shed lights on its difficulty in Section~\ref{sGeneralized}. 
In Section~\ref{sApproach}, we propose a method to remedy the observed issues in the previous section and compare it to related approaches. 
Experimental results, detailed analysis, and discussions are provided in Section~\ref{sExp},~\ref{sAnalysis}, and~\ref{sDiscuss}, respectively.

\section{Related Work}
\label{sRelated}

There has been very little work on generalized zero-shot learning.
\cite{FromeCSBDRM13,NorouziMBSSFCD14,MensinkVPC12,Tang2010optimizing} allow the label space of their classifiers to include seen classes but they only test on the data from the unseen classes.
\cite{SocherGMN13} proposes a two-stage approach that first determines whether a test data point is from a seen or unseen class, and then apply the corresponding classifiers. However, their experiments are limited to only 2 or 6 unseen classes. We describe and compare to their methods in Section~\ref{sNovelty}, ~\ref{sExp}, and the Supplementary Material.
In the domain of action recognition, \cite{gan2016recognizing} investigates the generalized setting with only up to 3 seen classes.
\cite{ElhoseinySE13} and \cite{LeiSFS15} focus on training a zero-shot binary classifier for \emph{each} unseen class (against seen ones) --- it is not clear how to distinguish multiple unseen classes from the seen ones. Finally, open set recognition~\cite{scheirer2013toward,scheirer2014probability,jain2014multi} considers testing on both types of classes, but treating the unseen ones as a single outlier class.

\section{Generalized Zero-Shot Learning}
\label{sGeneralized}

In this section, we describe formally the setting of \textit{generalized zero-shot learning}. We then present empirical evidence to illustrate the difficulty of this problem.

\subsection{Conventional and Generalized Zero-shot Learning}

Suppose we are given the training data $\mathcal{D}= \{(\vx_n\in \R^{\cD},y_n)\}_{n=1}^\cN$ with the labels $y_n$  from the label space of \emph{seen} classes $\mathcal{S} = \{1,2,\cdots,\cS\}$.  Denote by $\mathcal{U} = \{\cS+1,\cdots,\cS+\cU\}$ the label space of \emph{unseen} classes. We use $\mathcal{T} = \mathcal{S} \cup \mathcal{U}$ to represent the union of the two sets of classes.

In the (conventional) zero-shot learning (ZSL) setting, the main goal is to classify test data into the \textit{unseen} classes, assuming the absence of the seen classes in the test phase.
In other words, each test data point is assumed to come from and will be assigned to one of the labels in $\mathcal{U}$.

Existing research on ZSL has been almost entirely focusing on this setting \cite{LampertNH09,PalatucciPHM09,YuA10,RohrbachSS11,KankuekulKTH12,AkataPHS13,YuCFSC13,FromeCSBDRM13,MensinkGS14,NorouziMBSSFCD14,JayaramanG14,AlHalahS15,AkataRWLS15,FuHXG15,FuXKG15,LiGS15,Bernardino15,KodirovXFG15,ZhangS15,zhang2016classifying,ChangpinyoCGS16}. However, in real applications, the assumption of encountering data only from the unseen classes is hardly realistic. The seen classes are often the most common objects we see in the real world. Thus, the objective in the conventional ZSL does not truly reflect how the classifiers will perform recognition in the wild.

Motivated by this shortcoming of the conventional ZSL, we advocate studying the more general setting of~\textit{generalized zero-shot learning (GZSL)}, where we no longer limit the possible class memberships of test data ---  each of them belongs to one of the classes in $\mathcal{T}$.

\subsection{Classifiers}

Without the loss of generality, we assume that for each class $c \in\mathcal{T}$, we have a discriminant scoring function $f_c(\vx)$, from which we would be able to derive the label for $\vx$.  For instance, for an unseen class $u$, the method of synthesized classifiers~\cite{ChangpinyoCGS16} defines $f_u(\vx) = \vw_u\T\vx$, where $\vw_u$ is the model parameter vector for the class $u$, constructed from its semantic embedding $\va_u$ (such as its attribute vector or the word vector associated with the name of the class). In ConSE \cite{NorouziMBSSFCD14}, $f_u(\vx) = \cos(s(\vx), \va_u)$, where $s(\vx)$ is the predicted embedding of the data sample $\vx$.
In DAP/IAP \cite{LampertNH14}, $f_u(\vx)$ is a probabilistic model of attribute vectors. We assume that similar discriminant functions for seen classes can be constructed in the same manner given their corresponding semantic embeddings.

How to assess an algorithm for GZSL?  We define and differentiate the following performance metrics: $A_{\mathcal{U} \rightarrow \mathcal{U}}$  the accuracy of classifying test data from $\mathcal{U}$ into $\mathcal{U}$, $A_{\mathcal{S} \rightarrow \mathcal{S}}$ the accuracy of classifying test data from $\mathcal{S}$ into $\mathcal{S}$, and finally  $A_{\mathcal{S} \rightarrow \mathcal{T}}$ and $A_{\mathcal{U} \rightarrow \mathcal{T}}$ the accuracies of classifying test data from either seen or unseen classes into the joint labeling space. Note that $A_{\mathcal{U} \rightarrow \mathcal{U}}$ is the standard performance metric used for conventional ZSL and $A_{\mathcal{S} \rightarrow \mathcal{S}}$ is the standard metric for multi-class classification. Furthermore, note that we do not report $A_{\mathcal{T} \rightarrow \mathcal{T}}$ as simply averaging $A_{\mathcal{S} \rightarrow \mathcal{T}}$ and $A_{\mathcal{U} \rightarrow \mathcal{S}}$ to compute $A_{\mathcal{T} \rightarrow \mathcal{T}}$  might be misleading when the two metrics are not balanced, as shown below.

\subsection{Generalized ZSL is hard}

To demonstrate the difficulty of GZSL, we report the empirical results of using a simple but intuitive algorithm for GZSL.  Given the discriminant functions, we adopt the following classification rule
\begin{align}
\hat{y} = \arg\max_{c\in\mathcal{T}} \quad f_c(\vx) 
\label{ePredict}
\end{align}
which we refer to as \emph{direct stacking}. 

We use the rule on ``stacking'' classifiers from the following zero-shot learning approaches: DAP and IAP \cite{LampertNH14}, ConSE \cite{NorouziMBSSFCD14}, and Synthesized Classifiers (SynC) \cite{ChangpinyoCGS16}.
We tune the hyper-parameters for each approach based on class-wise cross validation \cite{ChangpinyoCGS16,ZhangS15,ElhoseinySE13}.   We test GZSL on two datasets \textbf{AwA}~\cite{LampertNH14} and \textbf{CUB}~\cite{WahCUB_200_2011} --- details about those datasets can be found in Section~\ref{sExp}.

\begin{table}[t]
\centering
{\small
\caption{\small Classification accuracies (\%) on conventional ZSL ($A_{\mathcal{U} \rightarrow \mathcal{U}}$), multi-class classification for seen classes ($A_{\mathcal{S} \rightarrow \mathcal{S}}$), and GZSL ($A_{\mathcal{S} \rightarrow \mathcal{T}}$ and $A_{\mathcal{U} \rightarrow \mathcal{T}}$), on \textbf{AwA} and \textbf{CUB}. Significant drops are observed from $A_{\mathcal{U} \rightarrow \mathcal{U}}$ to $A_{\mathcal{U} \rightarrow \mathcal{T}}$.}
\label{ana}
\begin{tabular}{c|c|c|c|c||c|c|c|c} \hline
& \multicolumn{4}{|c||}{\textbf{AwA}} & \multicolumn{4}{|c}{\textbf{CUB}} \\ \cline{2-9}
Method & $A_{\mathcal{U} \rightarrow \mathcal{U}}$ & $A_{\mathcal{S} \rightarrow \mathcal{S}}$ & $A_{\mathcal{U} \rightarrow \mathcal{T}}$ & $A_{\mathcal{S} \rightarrow \mathcal{T}}$ 
& $A_{\mathcal{U} \rightarrow \mathcal{U}}$ & $A_{\mathcal{S} \rightarrow \mathcal{S}}$ & $A_{\mathcal{U} \rightarrow \mathcal{T}}$ & $A_{\mathcal{S} \rightarrow \mathcal{T}}$ \\ \hline
DAP~\cite{LampertNH14} & 51.1& 78.5& 2.4& 77.9 & 38.8& 56.0& 4.0& 55.1\\
IAP~\cite{LampertNH14} & 56.3& 77.3& 1.7& 76.8 & 36.5& 69.6& 1.0& 69.4\\
ConSE~\cite{NorouziMBSSFCD14} & 63.7& 76.9& 9.5& 75.9 & 35.8& 70.5& 1.8& 69.9\\
{SynC$^\textrm{o-vs-o}$}~\cite{ChangpinyoCGS16} & 70.1& 67.3& 0.3& 67.3 & 53.0& 67.2& 8.4& 66.5\\
{SynC$^\textrm{struct}$}~\cite{ChangpinyoCGS16} & 73.4& 81.0& 0.4& 81.0 & 54.4& 73.0& 13.2& 72.0\\ \hline
\end{tabular}
}
\end{table}

Table~\ref{ana} reports experimental results based on the 4 performance metrics we have described previously. Our goal here is \text{\emph{not}} to compare between methods. Instead, we examine the impact of relaxing the assumption of  \emph{the prior knowledge of} whether  data are from seen or unseen classes.

We observe that, in this setting of GZSL, the classification performance for unseen classes ($A_{\mathcal{U} \rightarrow \mathcal{T}}$) drops significantly from the performance in conventional ZSL ($A_{\mathcal{U} \rightarrow \mathcal{U}}$), while that of seen ones ($A_{\mathcal{S} \rightarrow \mathcal{T}}$) remains roughly the same as in the multi-class task  ($A_{\mathcal{S} \rightarrow \mathcal{S}}$). That is, \emph{nearly all test data from unseen classes are misclassified into the seen classes}. This unusual degradation in performance highlights the challenges of GZSL; as we only see labeled data from seen classes during training, the scoring functions of seen classes tend to dominate those of unseen classes, leading to biased predictions in GZSL and  aggressively classifying a new data point into the label space of $\mathcal{S}$ because classifiers for the seen classes do not get trained on ``negative'' examples from the unseen classes.

\section{Approach for GZSL}
\label{sApproach}

The previous example shows that the classifiers for unseen classes constructed by conventional ZSL methods should not be naively combined with models for seen classes to expand the labeling space required by GZSL. 

In what follows, we propose a simple variant to the naive approach of \emph{direct stacking} to curb such a problem. We also develop a metric that measures the performance of GZSL, by acknowledging that there is an inherent trade-off between recognizing seen classes and recognizing unseen classes. This metric, referred to as the Area Under Seen-Unseen accuracy Curve (AUSUC), balances the two conflicting forces.  We conclude this section by describing two related approaches: despite their sophistication, they do not perform well empirically.

\subsection{Calibrated stacking}
\label{sCalibrate}

Our approach stems from the observation that the scores of the discriminant functions for the seen classes are often greater than the scores for the unseen classes. Thus, intuitively, we would like to reduce the scores for the seen classes. This leads to the following classification rule:
\begin{align}
\hat{y} = \arg\max_{c\;\in\;\mathcal{T}}\quad f_c(\vx) - \gamma\I[c\in\mathcal{S}],
\label{ePredictCalib}
\end{align}
where the indicator $\I[\cdot]\in\{0,1\}$ indicates whether or not $c$ is a seen class and $\gamma$ is a calibration factor. We term this adjustable rule as~\textit{calibrated stacking}.

Another way to interpret $\gamma$ is to regard it as the prior likelihood of a data point coming from unseen classes.  When $\gamma=0$, the calibrated stacking rule reverts back to the direct stacking rule, described previously. 

It is also instructive to consider the two extreme cases of $\gamma$. When $\gamma \rightarrow +\infty$, the classification rule will ignore all seen classes and classify all data points into one of the unseen classes. When there is no new data point coming from seen classes, this classification rule essentially implements what one would do in the setting of conventional ZSL. On the other hand, when $\gamma \rightarrow -\infty$, the classification rule only considers the label space of seen classes as in standard multi-way classification. 
The calibrated stacking rule thus represents a middle ground between aggressively classifying every data point into seen classes and conservatively classifying every data point into unseen classes. Adjusting this hyperparameter thus gives a trade-off, which we exploit to define a new performance metric. 

\subsection{Area Under Seen-Unseen Accuracy Curve (AUSUC)}

\begin{figure}[t]
\centering
\small
\includegraphics[width=0.4\textwidth]{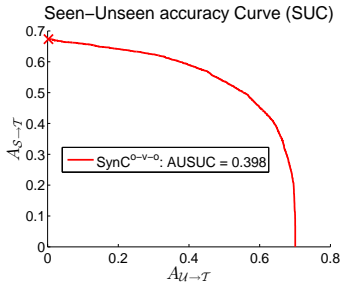}
\caption{\small The Seen-Unseen accuracy Curve (SUC) obtained by varying $\gamma$ in the calibrated stacking classification rule eq.~(\ref{ePredictCalib}). The AUSUC summarizes the curve by computing the area under it. We use the method SynC$^\textrm{o-vs-o}$ on the \textbf{AwA} dataset, and tune hyper-parameters as in Table~\ref{ana}. The red cross denotes the accuracies by direct stacking.}
\label{fSUC}
\end{figure} 

Varying the calibration factor $\gamma$, we can compute a series of classification accuracies ($A_{\mathcal{U} \rightarrow \mathcal{T}}$, $A_{\mathcal{S} \rightarrow \mathcal{T}}$). Fig.~\ref{fSUC} plots those points for the dataset \textbf{AwA} using the classifiers generated by the method in \cite{ChangpinyoCGS16} based on class-wise cross validation.  We call such a plot the \emph{Seen-Unseen accuracy Curve (SUC)}.

On the curve, $\gamma=0$ corresponds to direct stacking, denoted by a cross. The curve is similar to many familiar curves  for representing conflicting goals, such as the Precision-Recall (PR) curve and the Receiving Operator Characteristic (ROC) curve, with two ends for the extreme cases ($\gamma \rightarrow -\infty$ and $\gamma \rightarrow +\infty$). 

A convenient way to summarize the plot with one number is to use the Area Under SUC (AUSUC)\footnote{If a single $\gamma$ is desired, the ``F-score'' that balances $A_{\mathcal{U} \rightarrow \mathcal{T}}$ and $A_{\mathcal{S} \rightarrow \mathcal{T}}$ can be used.}. The higher the area is, the better an algorithm is able to balance  $A_{\mathcal{U} \rightarrow \mathcal{T}}$ and $A_{\mathcal{S} \rightarrow \mathcal{T}}$. In Section~\ref{sExp}, Section~\ref{sAnalysis}, and the Supplementary Material, we evaluate the performance of existing zero-shot learning methods under this metric, as well as provide further insights and analyses.

An immediate and important use of the metric AUSUC is for model selection. Many ZSL learning methods require tuning hyperparameters --- previous work tune them based on the accuracy $A_{\mathcal{U} \rightarrow \mathcal{U}}$. The selected model, however, does not necessarily balance optimally between $A_{\mathcal{U} \rightarrow \mathcal{T}}$ and $A_{\mathcal{S} \rightarrow \mathcal{T}}$. Instead, we advocate using AUSUC for model selection and hyperparamter tuning. Models with higher values of AUSUC are likely to perform in balance for the task of GZSL. For detailed discussions, see the Supplementary Material.

\subsection{Alternative approaches}
\label{sNovelty}

Socher et al. \cite{SocherGMN13} propose a two-stage zero-shot learning approach that first predicts whether an image is of seen or unseen classes and then accordingly applies the corresponding classifiers.  The first stage is based on the idea of novelty detection and assigns a high novelty  score if it is unlikely for the data point to come from seen classes. 
They experiment with two novelty detection strategies: Gaussian and LoOP models~\cite{KriegelKSZ09}. We briefly describe and contrast them to our approach below. The details are in the Supplementary Material.

\paragraph{\textbf{Novelty detection}} The main idea is to assign a novelty score $N(\vx)$ to each sample $\vx$.  With this novelty score, the final prediction rule becomes
\begin{align}
\hat{y} = \left\{
				\begin{array}{lr}
        \arg\max_{c\;\in\;\mathcal{S}}\quad f_c(\vx), \quad & \textrm{if } N(\vx) \leq -\gamma.\\
        \arg\max_{c\;\in\;\mathcal{U}}\quad f_c(\vx), \quad & \textrm{if } N(\vx) > -\gamma.
        \end{array}
				\right.
\label{ePredictNovelty}
\end{align}
where $-\gamma$ is the novelty threshold. The scores above this threshold indicate belonging to unseen classes. To estimate $N(\vx)$, for the Gaussian model, data points in seen classes are first modeled with a Gaussian mixture model. The novelty score of a data point is then its negative log probability value under this mixture model.  Alternatively, the novelty score can be estimated using the Local Outlier Probabilities (LoOP) model~\cite{KriegelKSZ09}.  The idea there is to compute the distances of $\vx$ to its nearest seen classes. Such distances are then converted to an outlier probability, interpreted as the likelihood of $\vx$ being from unseen classes.

\paragraph{\textbf{Relation to calibrated stacking}}  If we define a new form of novelty score $N(\vx) = \max_{u\;\in\;\mathcal{U}} f_u(\vx) - \max_{s\;\in\;\mathcal{S}} f_s(\vx)$ in eq.~(\ref{ePredictNovelty}), we recover the prediction rule in eq.~(\ref{ePredictCalib}). However, this relation holds only if we are interested in predicting one label $\hat{y}$. When we are interested in predicting a set of labels (for example, hoping that the correct labels are in the top $K$ predicted labels, (i.e., the Flat hit@K metric, cf. Section~\ref{sExp}), the two prediction rules will give different results.

\section{Experimental Results}
\label{sExp}

\subsection{Setup}

\paragraph{\textbf{Datasets}} We mainly use three benchmark datasets: the \textbf{Animals with Attributes (AwA)}~\cite{LampertNH14},  \textbf{CUB-200-2011 Birds (CUB)}~\cite{WahCUB_200_2011}, and \textbf{ImageNet} (with full 21,841 classes)~\cite{ILSVRC15}. Table~\ref{tDatasets} summarizes their key characteristics.

\begin{table}[t]
\centering
{\small
\caption{Key characteristics of the studied datasets.}
\label{tDatasets}
\begin{tabular}{c|c|c|c}\hline
Dataset &  Number of & Number of  & Total number \\ 
name & seen classes & unseen classes & of images\\ \hline
AwA$^\dagger$ & 40  & 10 & 30,475\\ \hline
CUB$^\ddagger$ & 150 & 50 & 11,788\\ \hline
ImageNet$^\S$ & 1000 & 20,842 & 14,197,122\\ \hline
\end{tabular}
\begin{flushleft}
{\scriptsize $^\dagger$: following the split in~\cite{LampertNH14}. $^\ddagger$: following~\cite{ChangpinyoCGS16} to report the average over 4 random splits. $^\S$: seen and unseen classes from ImageNet ILSVRC 2012 1K~\cite{ILSVRC15} and Fall 2011 release \cite{deng2009imagenet}, respectively.}
\end{flushleft}
}
\end{table}

\paragraph{\textbf{Semantic spaces}} For the classes in \textbf{AwA} and \textbf{CUB}, we use 85-dimensional and 312-dimensional binary or continuous-valued attributes, respectively~\cite{LampertNH14,WahCUB_200_2011}. For \textbf{ImageNet}, we use 500-dimensional word vectors (\textsc{word2vec}) trained by the skip-gram model~\cite{MikolovCCD13,MikolovSCCD13} provided by Changpinyo et al. \cite{ChangpinyoCGS16}. We ignore classes without word vectors, resulting in 20,345 (out of 20,842) unseen classes. We follow \cite{ChangpinyoCGS16} to normalize all but binary embeddings to have unit $\ell_2$ norms.

\paragraph {\textbf{Visual features}} We use the GoogLeNet deep features~\cite{SzegedyLJSRAEVR14} pre-trained on ILSVRC 2012 1K~\cite{ILSVRC15} for all datasets (all extracted with the Caffe package~\cite{jia2014caffe}). Extracted features come from the 1,024-dimensional activations of the pooling units, as in~\cite{AkataRWLS15,ChangpinyoCGS16}.

\paragraph{\textbf{Zero-shot learning methods}}
We examine several representative conventional zero-shot learning approaches, described briefly below.
Direct Attribute Prediction (DAP) and Indirect Attribute Prediction (IAP) \cite{LampertNH14} are probabilistic models that perform attribute predictions as an intermediate step and then use them to compute MAP predictions of unseen class labels. ConSE \cite{NorouziMBSSFCD14} makes use of pre-trained classifiers for seen classes and their probabilitic outputs to infer the semantic embeddings of each test example, and then classifies it into the unseen class with the most similar semantic embedding.
SynC \cite{ChangpinyoCGS16} is a recently proposed multi-task learning approach that synthesizes a novel classifier based on semantic embeddings and base classifiers that are learned with labeled data from the seen classes. Two versions of this approach --- SynC$^\textrm{o-v-o}$ and SynC$^\textrm{struct}$ --- use one-versus-other and Crammer-Singer style~\cite{CrammerS02} loss functions to train classifiers. We use binary attributes for DAP and IAP, and continuous attributes and \textsc{word2vec} for ConSE and SynC, following \cite{LampertNH14,NorouziMBSSFCD14,ChangpinyoCGS16}.

\paragraph{\textbf{Generalized zero-shot learning tasks}}
There are no previously established benchmark tasks for GZSL. We thus define a set of tasks that reflects more closely how data are distributed in real-world applications. 

We construct the GZSL tasks by composing test data as a combination of images from both seen and unseen classes. 
We follow existing splits of the datasets for the conventional ZSL to separate seen and unseen classes. 
Moreover, for the datasets \textbf{AwA} and \textbf{CUB}, we hold out 20$\%$ of the data points from the seen classes (previously, all of them are used for training in the conventional zero-shot setting) and merge them with the data from the unseen classes to form the test set; for \textbf{ImageNet}, we combine its validation set (having the same classes as its training set) and the 21K classes that are not in the ILSVRC 2012 1K dataset.

\paragraph{\textbf{Evaluation metrics}} While we will primarily report the performance of ZSL approaches under the metric Area Under Seen-Unseen accuracy Curve (AUSUC) developed in Section~\ref{sCalibrate}, we explain how its two accuracy components $A_{\mathcal{S} \rightarrow \mathcal{T}}$ and $A_{\mathcal{U} \rightarrow \mathcal{T}}$ are computed below.

For \textbf{AwA} and \textbf{CUB}, seen and unseen accuracies correspond to (normalized-by-class-size) multi-way classification accuracy, where the seen accuracy is computed on the 20$\%$ images from the seen classes and the unseen accuracy is computed on images from unseen classes.

For \textbf{ImageNet}, seen and unseen accuracies correspond to Flat hit@K (F@K), defined as the percentage of test images for which the model returns the true label in its top K predictions. Note that, F@1 is the unnormalized multi-way classification accuracy. Moreover, following the procedure in \cite{FromeCSBDRM13,NorouziMBSSFCD14,ChangpinyoCGS16}, we evaluate on three scenarios of increasing difficulty:
(1)  \emph{2-hop} contains 1,509 unseen classes that are within two tree hops of the 1K seen classes according to the ImageNet label hierarchy\footnote{\url{http://www.image-net.org/api/xml/structure_released.xml}}.
(2) \emph{3-hop} contains 7,678 unseen classes that are within three tree hops of the seen classes.
(3) \emph{All} contains all 20,345 unseen classes.

\subsection{Which method to use to perform GZSL?}

Table~\ref{tNov} provides an experimental comparison between several  methods utilizing seen and unseen classifiers for generalized ZSL,  with hyperparameters cross-validated to maximize AUSUC. Empirical results on additional datasets and ZSL methods are in the Supplementary Material. 

The results show that, irrespective of which ZSL methods are used to generate models for seen and unseen classes, our method of \emph{calibrated stacking} for generalized ZSL outperforms other methods. In particular, despite their probabilistic justification, the two novelty detection methods do not perform well. We believe that this is because most existing zero-shot learning methods are discriminative and optimized to  take full advantage of class labels and semantic information. In contrast, either Gaussian or LoOP approach models all the seen classes as a whole, possibly at the cost of modeling inter-class differences.

\begin{table}
\centering
{\small
\caption{Performances measured in AUSUC of several methods for Generalized Zero-Shot Learning on \textbf{AwA} and \textbf{CUB}. The higher the better (the upper bound is 1).}
\label{tNov}
\begin{tabular}{c|c|c|c||c|c|c} \hline
& \multicolumn{3}{|c||}{\textbf{AwA}} & \multicolumn{3}{|c}{\textbf{CUB}} \\ \cline{2-7}
Method  & \multicolumn{2}{|c|}{Novelty detection \cite{SocherGMN13}} & Calibrated  &  \multicolumn{2}{|c|}{Novelty detection \cite{SocherGMN13}} & Calibrated \\ \cline{2-3} \cline{5-6} 
 & Gaussian & LoOP & Stacking & Gaussian & LoOP & Stacking \\ 
\hline
DAP & 0.302 & 0.272 & 0.366 & 0.122 & 0.137 & 0.194\\
IAP & 0.307 & 0.287 & 0.394 & 0.129 & 0.145 & 0.199\\
ConSE & 0.342 & 0.300 & 0.428 & 0.130 & 0.136 & 0.212\\
{SynC$^\textrm{o-vs-o}$} & 0.420 &  0.378 & 0.568 & 0.191 & 0.209 & 0.336\\
{SynC$^\textrm{struct}$} & 0.424 &  0.373 & 0.583 & 0.199 & 0.224 & 0.356\\ \hline
\end{tabular}
}
\end{table}

\begin{figure}[t]
\centering
\includegraphics[width=.45\textwidth]{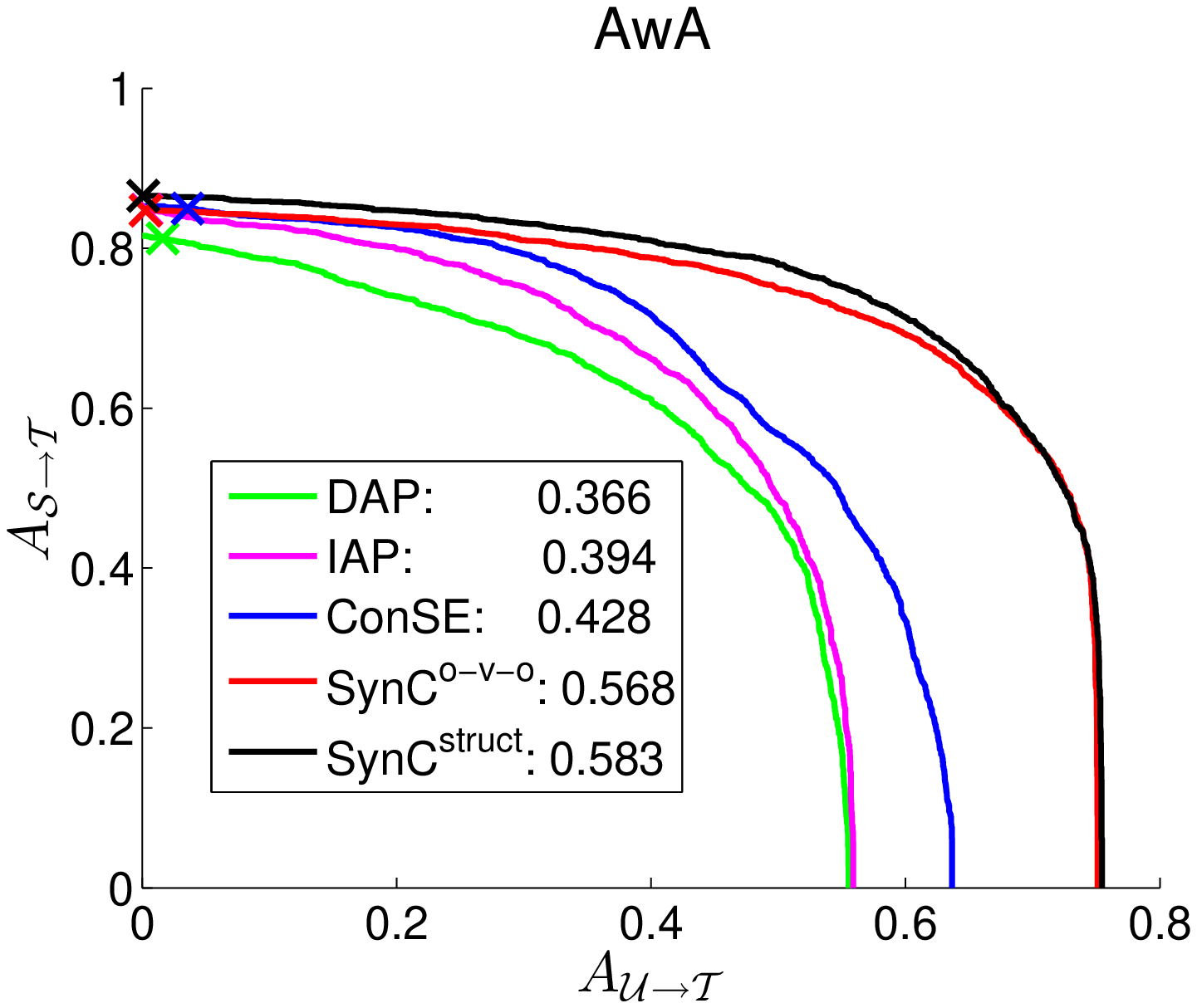}\quad\includegraphics[width=.45\textwidth]{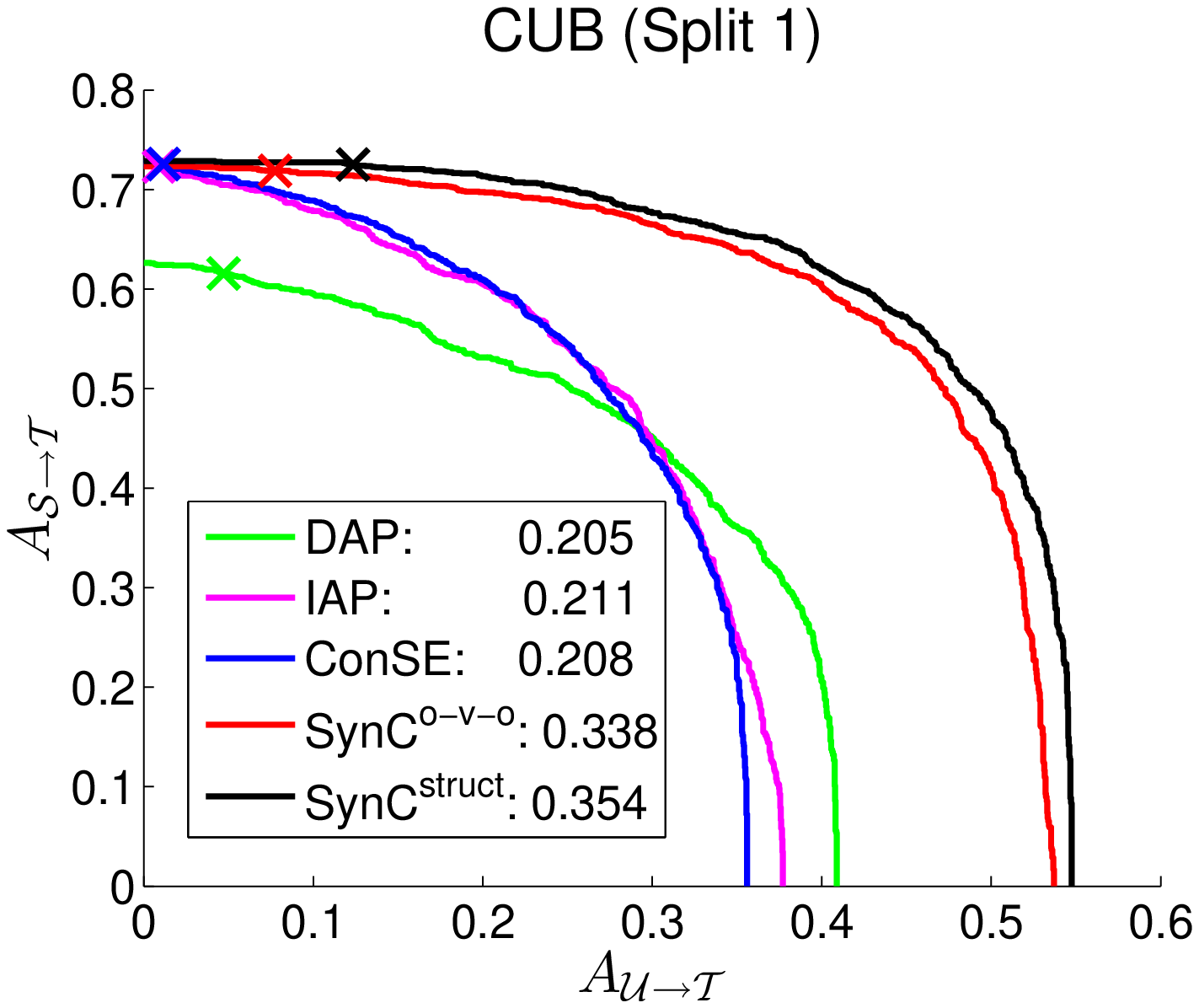}
\caption{\small Comparison between several ZSL approaches on the task of GZSL for \textbf{AwA} and \textbf{CUB}.}
\label{fAUSUC_AwA_CUB}
\end{figure}

\begin{figure}[!ht]
\centering
\includegraphics[width=.39\textwidth]{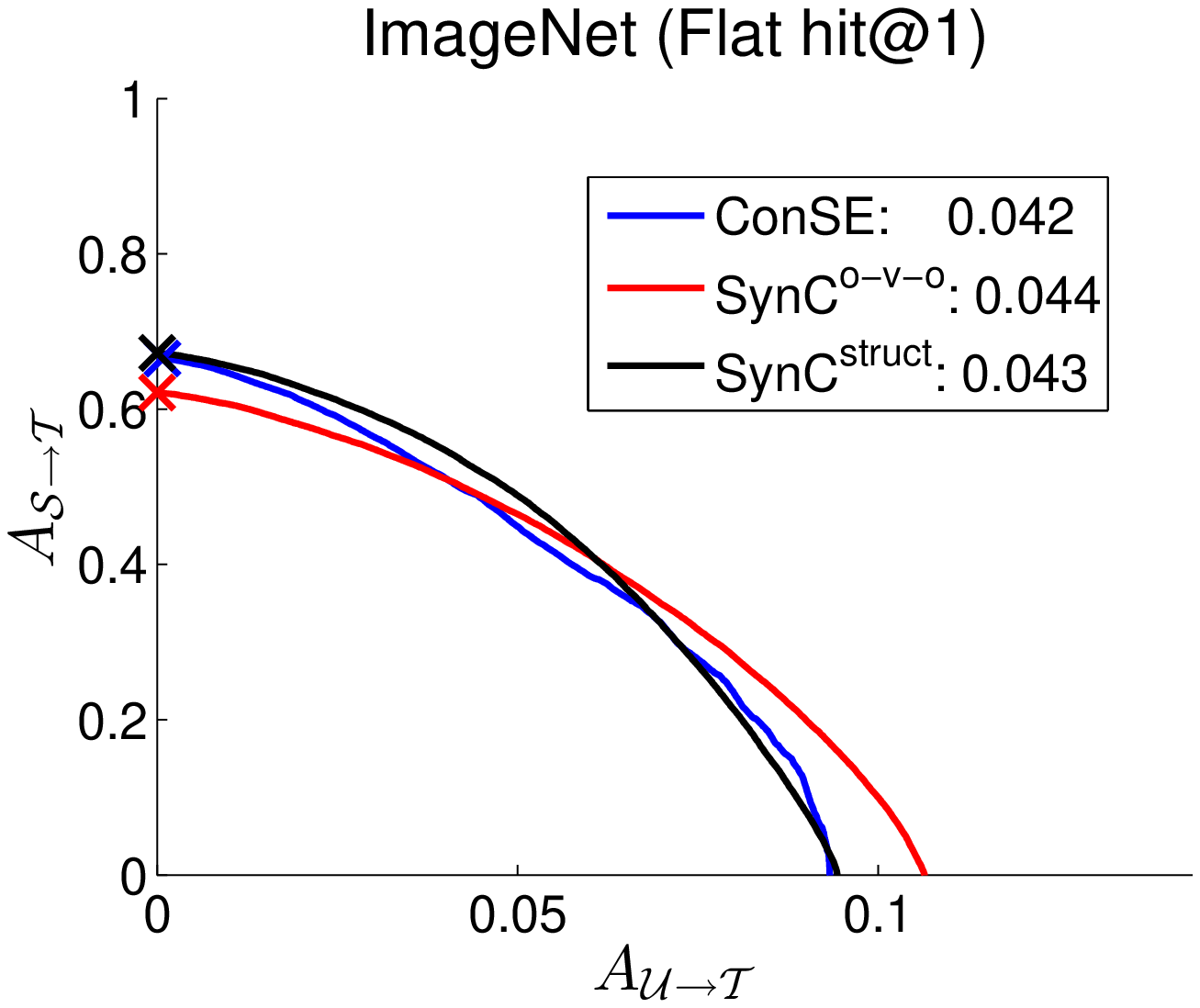}
\includegraphics[width=.39\textwidth]{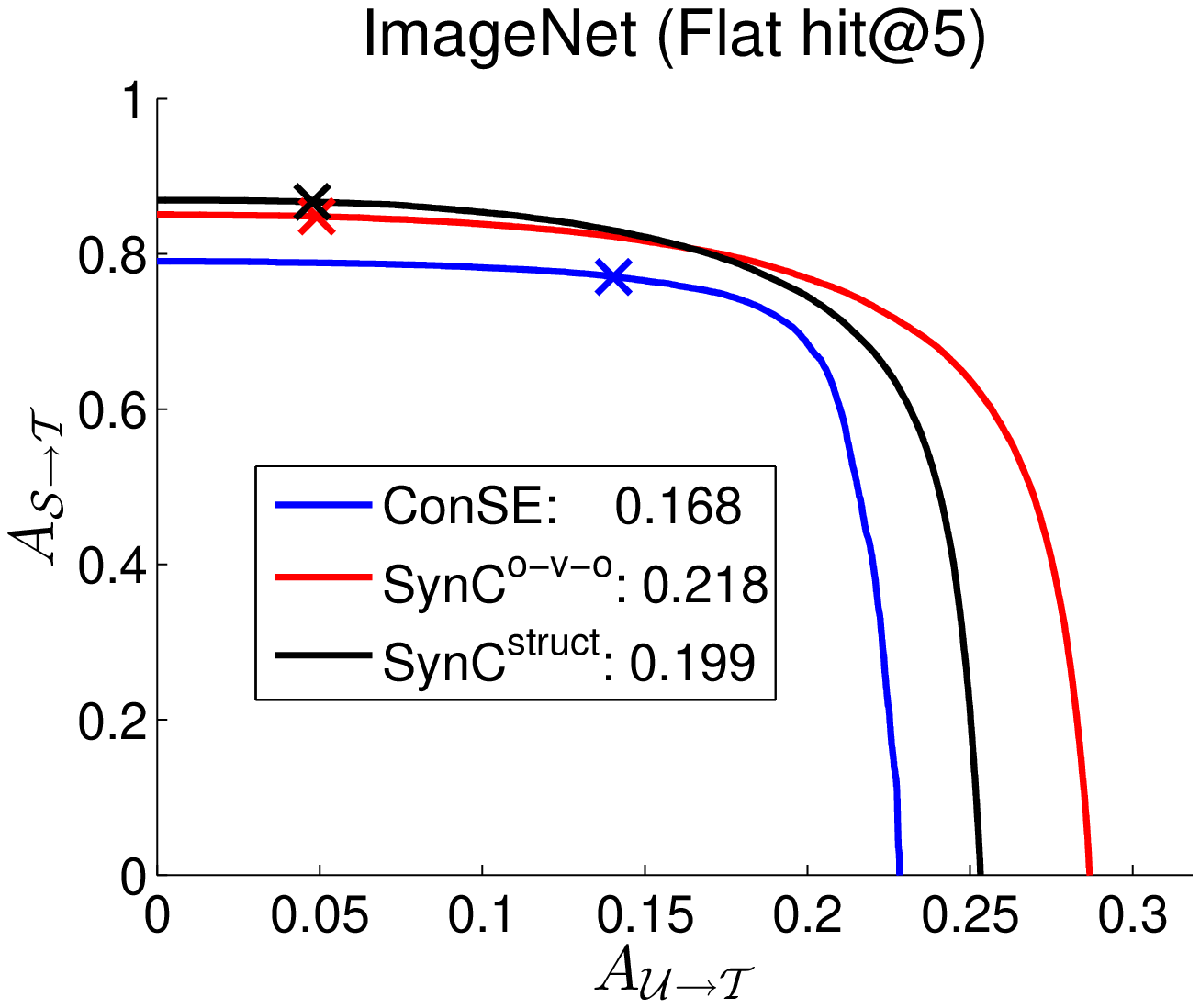}
\includegraphics[width=.39\textwidth]{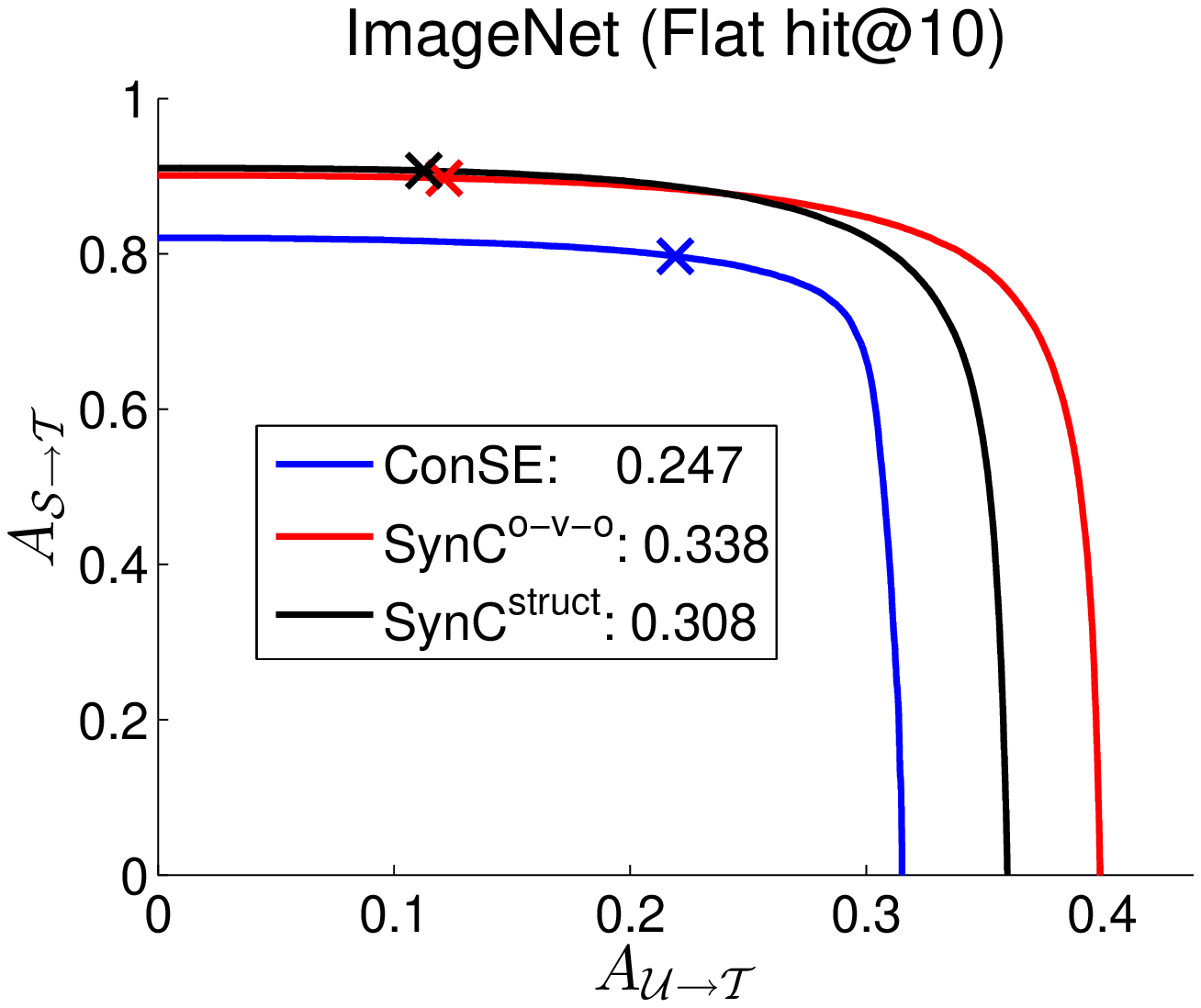}
\includegraphics[width=.39\textwidth]{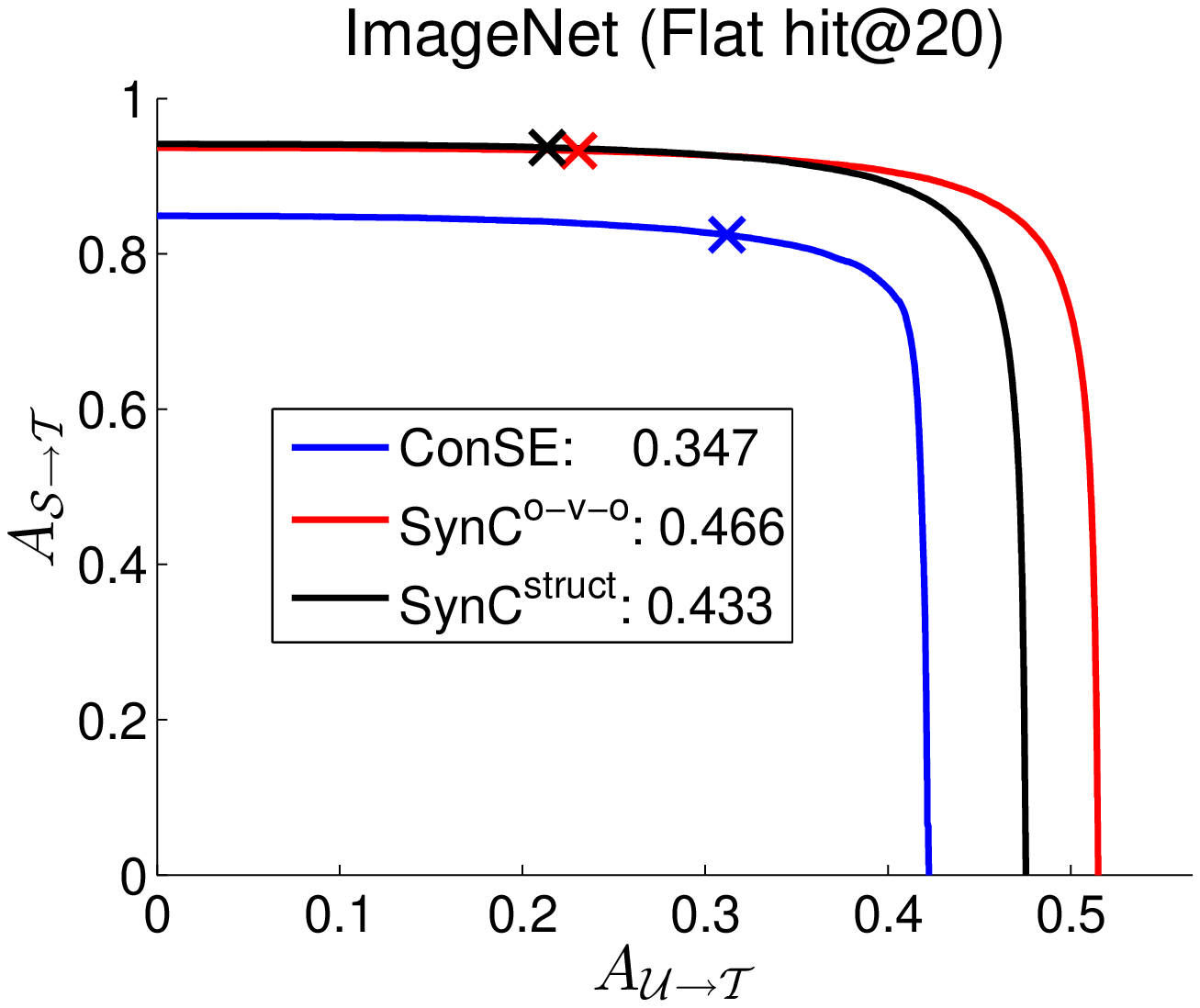}
\caption{\small Comparison between ConSE and SynC of their performances on the task of GZSL for \textbf{ImageNet} where the unseen classes are within 2 tree-hops from seen classes.}
\label{fAUSUC_ImageNet_2hop}
\end{figure}

\begin{table}[t!]
\centering
{\small
\caption{Performances measured in AUSUC by different zero-shot learning approaches on GZSL on \textbf{ImageNet}, using our method of \emph{calibrated stacking}.}
\label{tAUSUC_ImageNet}
\begin{tabular}{c|c|c|c|c|c} \hline
\text{Unseen} & \text{Method} & \multicolumn{4}{|c}{Flat hit@K} \\ \cline{3-6}
classes & & \hspace{15pt}1\hspace{15pt} & \hspace{15pt}5\hspace{15pt} & \hspace{15pt}10\hspace{15pt} & \hspace{15pt}20\hspace{15pt} \\ \hline 
\emph{2-hop} & ConSE      & 0.042 & 0.168 & 0.247 & 0.347 \\ 
						 & SynC$^\textrm{o-vs-o}$ & 0.044 & 0.218 & 0.338 & 0.466\\
						 & SynC$^\textrm{struct}$ & 0.043 & 0.199 & 0.308 & 0.433\\
\hline
\emph{3-hop} & ConSE      & 0.013 & 0.057 & 0.090 & 0.135\\ 
						 & SynC$^\textrm{o-vs-o}$ & 0.012 & 0.070 & 0.119 & 0.186 \\
						 & SynC$^\textrm{struct}$ & 0.013 & 0.066 & 0.110 & 0.170 \\
\hline
\emph{All}   & ConSE      & 0.007 & 0.030 & 0.048 & 0.073 \\ 
						 & SynC$^\textrm{o-vs-o}$ & 0.006 & 0.034 & 0.059 & 0.097\\
						 & SynC$^\textrm{struct}$ & 0.007 & 0.033 & 0.056 & 0.090 \\
\hline
\end{tabular}
}
\end{table}

\subsection{Which Zero-shot Learning approach is more robust to GZSL?}

Fig.~\ref{fAUSUC_AwA_CUB} contrasts in detail several ZSL approaches when tested on the task of GZSL, using the method of \emph{calibrated stacking}. Clearly, the SynC method dominates all other methods in the whole ranges. The crosses on the plots mark the results of \emph{direct stacking} (Section~\ref{sGeneralized}). 

Fig.~\ref{fAUSUC_ImageNet_2hop} contrasts in detail ConSE to SynC, the two known methods for large-scale ZSL. When the accuracies measured in Flat hit@1 (i.e., multi-class classification accuracy), neither method dominates the other, suggesting the different trade-offs by the two methods. However, when we measure hit rates in the top $K>1$, SynC dominates ConSE.  
Table~\ref{tAUSUC_ImageNet} gives summarized comparison in AUSUC between the two methods on the \textbf{ImageNet} dataset. We observe that SynC in general outperforms ConSE except when Flat hit@1 is used, in which case the two methods' performances are nearly indistinguishable. Additional plots can be found in the Supplementary Material.

\section{Analysis on (Generalized) Zero-shot Learning}
\label{sAnalysis}

Zero-shot learning, either in conventional setting or generalized setting, is a challenging problem as there is no labeled data for the unseen classes. The performance of  ZSL methods depends on at least two factors:  (1) how seen and unseen classes are related; (2) how effectively the relation can be exploited by learning algorithms to generate models for the unseen classes. For generalized zero-shot learning, the performance further depends on how classifiers for seen and unseen classes are combined to classify new data into the joint label space.

Despite extensive study in ZSL, several questions remain understudied. For example, given a dataset and a split of seen and unseen classes, what is the best possible performance of any ZSL method? How far are we from there? What is the most crucial component we can improve in order to reduce the gap between the state-of-the-art and the ideal performances?

In this section, we empirically analyze ZSL methods in detail and shed light on some of those questions.

\paragraph{\textbf{Setup}} As ZSL methods do not use labeled data from unseen classes for training classifiers, one reasonable estimate of their best possible performance is to measure the performance on a multi-class classification task where annotated data on the unseen classes are provided.

Concretely, to construct the multi-class classification task,  on \textbf{AwA} and \textbf{CUB}, we randomly select 80\% of the data along with their labels from all classes (seen and unseen) to train classifiers. The remaining 20\% will be used to assess both the multi-class classifiers and the classifiers from ZSL.  Note that, for ZSL, only the seen classes from the 80\% are used for training --- the portion belonging to the unseen classes are not used.   

On \textbf{ImageNet}, to reduce the computational cost (of constructing multi-class classifiers which would involve 20,345-way classification), we subsample another 1,000 unseen classes from its original 20,345 unseen classes.  We call this new dataset \textbf{ImageNet-2K} (including the 1K seen classes from \textbf{ImageNet}).  The subsampling procedure is described in the Supplementary Material and the main goal is to keep the proportions of difficult unseen classes unchanged.  Out of those 1,000 unseen classes, we randomly select 50 samples per class and reserve them for testing and use the remaining examples (along with their labels) to train 2000-way classifiers.

For ZSL methods, we use either attribute vectors or word vectors (\textsc{word2vec}) as semantic embeddings. Since SynC$^\textrm{o-vs-o}$~\cite{ChangpinyoCGS16} performs well on a range of datasets and settings, we focus on this method. For multi-class classification, we train one-versus-others SVMs. Once we obtain the classifiers for both seen and unseen classes, we use the \emph{calibrated stacking} decision rule to combine (as in generalized ZSL) and vary the calibration factor $\gamma$ to obtain the Seen-Unseen accuracy Curve, exemplified in Fig.~\ref{fSUC}.

\begin{figure}[t]
\centering
\includegraphics[width=.44\textwidth]{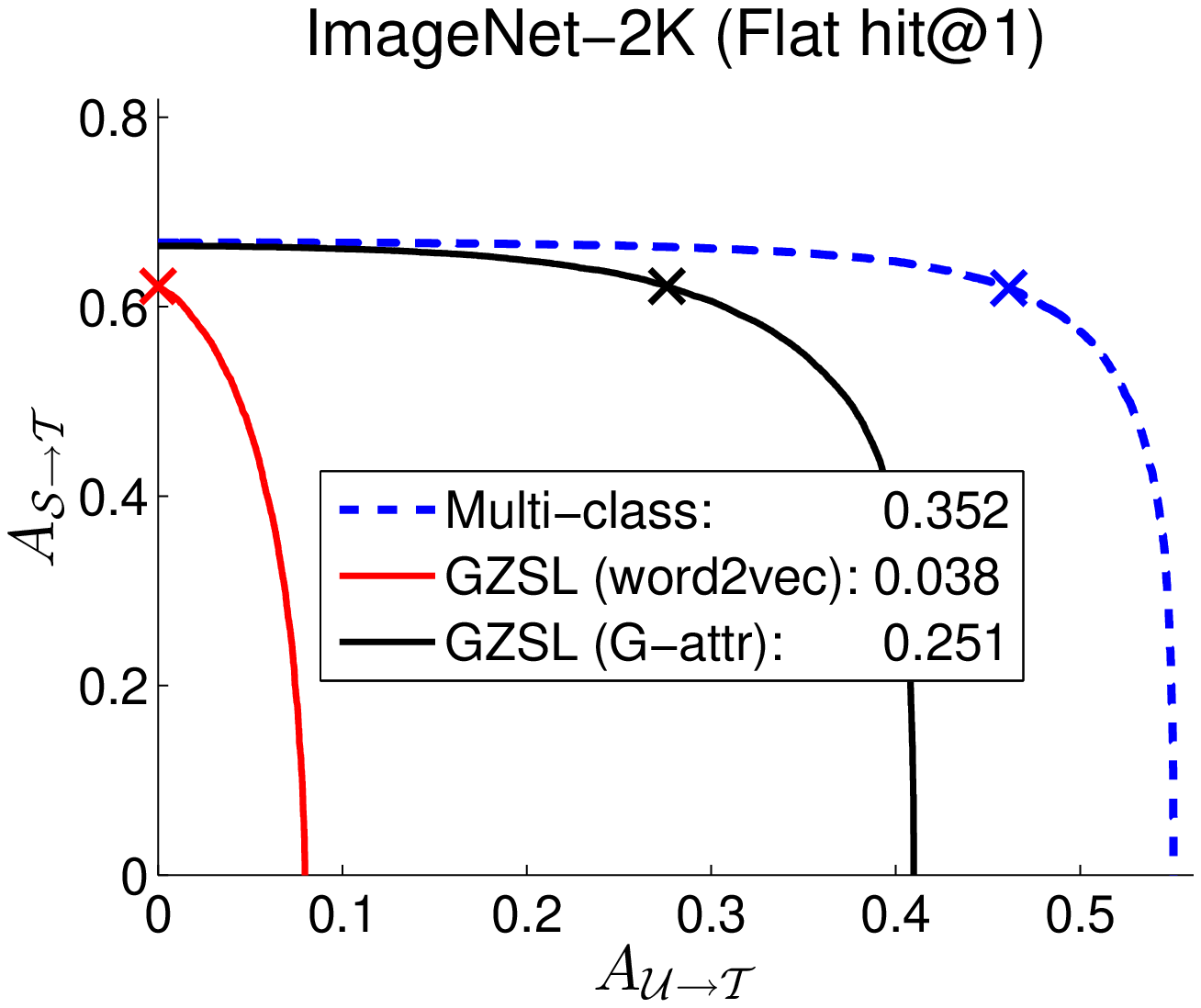}
\includegraphics[width=.44\textwidth]{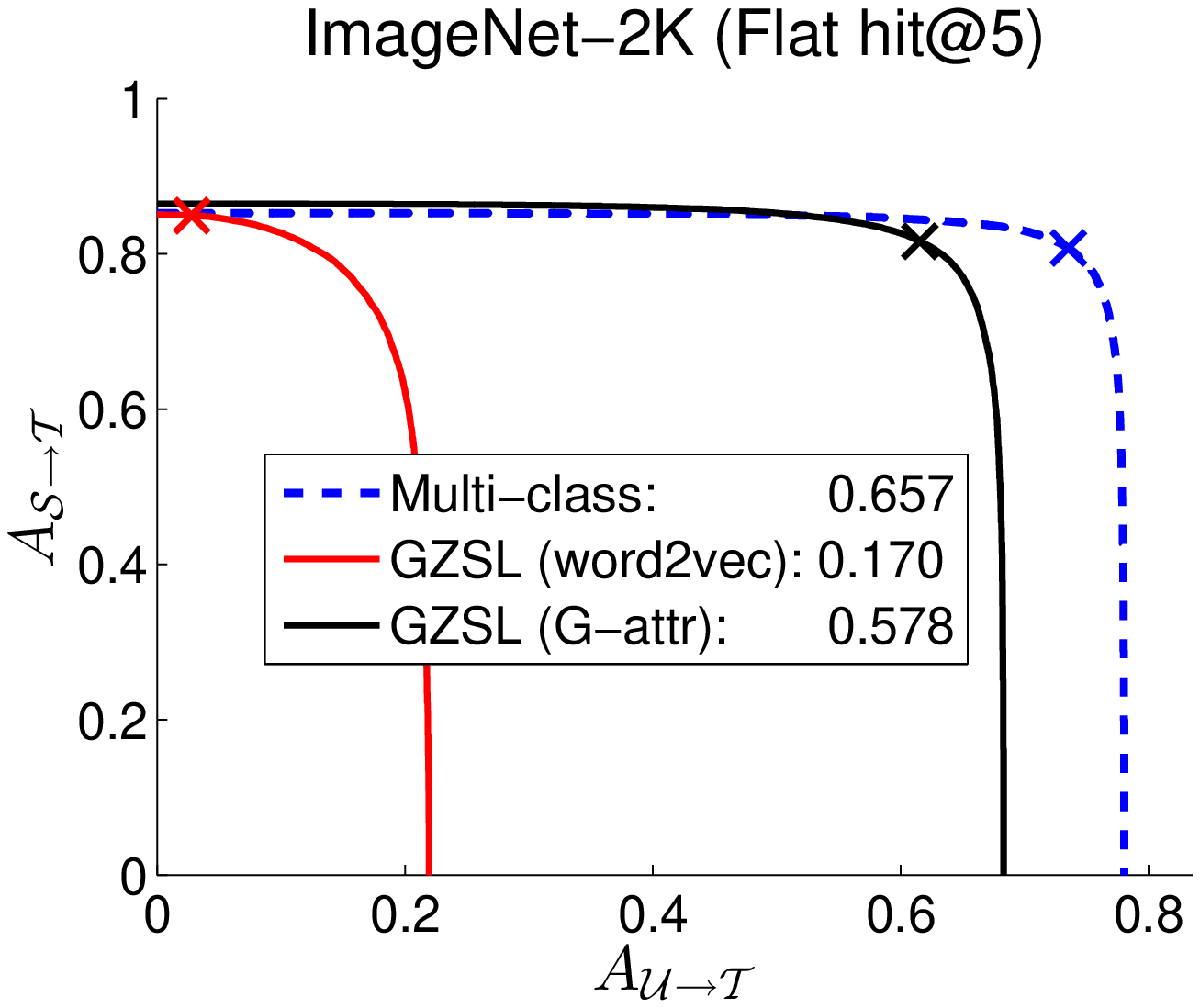}
\caption{\small We contrast the performances of GZSL to multi-class classifiers trained with labeled data from both seen and unseen classes on the dataset \textbf{ImageNet-2K}. GZSL uses \textsc{word2vector} (in red color) and the idealized visual features (G-attr) as semantic embeddings (in black color).}
\label{fig:ImageNet_new}
\end{figure}

\paragraph{\textbf{How far are we from the ideal performance}?}  Fig.~\ref{fig:ImageNet_new} displays the Seen-Unseen accuracy Curves for \textbf{ImageNet-2K} --- additional plots on \textbf{ImageNet-2K} and similar ones on \textbf{AwA} and \textbf{CUB} are in the Supplementary Material. Clearly, there is a large gap between the performances of GZSL using the default \textsc{word2vec} semantic embeddings and the ideal performance indicated by the multi-class classifiers. Note that the cross marks indicate the results of \emph{direct stacking}.  The multi-class classifiers not only dominate GZSL in the whole ranges (thus, with very high AUSUCs) but also are capable of learning classifiers that are well-balanced (such that \emph{direct stacking} works well).

\begin{table}[t]
\centering
{\small
\caption{Comparison of performances measured in AUSUC between GZSL (using \textsc{word2vec} and \textbf{G-attr}) and multi-class classification on \textbf{ImageNet-2K}. Few-shot results are averaged over 100 rounds. GZSL with \textbf{G-attr} improves upon GZSL with \textsc{word2vec} significantly and quickly approaches multi-class classification performance.}
\label{detail_ImageNet}
\begin{tabular}{c|l|c|c|c|c} \hline
\multicolumn{2}{c|}{Method} & \multicolumn{4}{|c}{Flat hit@K} \\ \cline{3-6}
\multicolumn{2}{c|}{}         & 1 & 5 & 10 & 20\\ \hline
& \multicolumn{1}{c|}{\textsc{word2vec}} & 0.04  & 0.17 & 0.27& 0.38\\ \cline{2-6}
& G-attr from 1 image  & 0.08\scriptsize{$\pm$0.003} & 0.25\scriptsize{$\pm$0.005} & 0.33\scriptsize{$\pm$0.005} & 0.42\scriptsize{$\pm$0.005}\\
GZSL & G-attr from 10 images & 0.20\scriptsize{$\pm$0.002} & 0.50\scriptsize{$\pm$0.002} & 0.62\scriptsize{$\pm$0.002} & 0.72\scriptsize{$\pm$0.002}\\
& G-attr from 100 images  & 0.25\scriptsize{$\pm$0.001} & 0.57\scriptsize{$\pm$0.001}& 0.69\scriptsize{$\pm$0.001} & 0.78\scriptsize{$\pm$0.001}\\
& G-attr from all images & 0.25 & 0.58& 0.69& 0.79\\  \hline
\multicolumn{2}{c|}{Multi-class classification} & 0.35 & 0.66 & 0.75& 0.82\\  \hline
\end{tabular}
}
\end{table}
\paragraph{\textbf{How much can idealized semantic embeddings help}?} We hypothesize that a large portion of the gap between GZSL and multi-class classification can be attributed to the weak semantic embeddings used by the GZSL approach.  

We investigate this by using a form of \emph{idealized} semantic embeddings. As the success of zero-shot learning  relies heavily on how accurate semantic embeddings represent visual similarity among classes, we examine the idea of \textit{visual features as semantic embeddings}.  Concretely, for each class, semantic embeddings can be obtained by averaging visual features of images belonging to that class. We call them \textbf{G-attr} as we derive the visual features from GoogLeNet. Note that, for unseen classes, we only use the reserved training examples to derive the semantic embeddings; we do not use their labels to train classifiers.

Fig.~\ref{fig:ImageNet_new} shows the performance of GZSL using \textbf{G-attr} --- the gaps to the multi-class classification performances are significantly reduced from those made by GZSL using \textsc{word2vec}. In some cases (see the Supplementary Material for more comprehensive experiments), GZSL can almost match the performance of multi-class classifiers without using any labels from the unseen classes!

\paragraph{\textbf{How much labeled data do we need to improve GZSL's performance}?}  Imagine we are given a budget to label data from unseen classes, how much those labels can improve GZSL's performance?  

Table~\ref{detail_ImageNet} contrasts the AUSUCs obtained by GZSL to those from mutli-class classification on \textbf{ImageNet-2K}, where GZSL is allowed to use visual features as embeddings --- those features can be computed from a few labeled images from the unseen classes, a scenario we can refer to as ``few-shot'' learning. Using about (randomly sampled) 100 labeled images per class, GZSL can quickly approach the performance of multi-class classifiers, which use about 1,000 labeled images per class. Moreover, those G-attr visual features as semantic embeddings improve upon \textsc{word2vec} more significantly under Flat hit@$\text{K}=1$ than when K $>$ 1.

We further examine on the whole~\textbf{ImageNet} with 20,345 unseen classes in Table~\ref{detail_ImageNet_all}, where we keep 80\% of the unseen classes' examples to derive \textbf{G-attr} and test on the rest, and observe similar trends. Specifically on Flat hit@1, the performance of G-attr from merely 1 image is boosted \textbf{threefold} of that by~\textsc{word2vec}, while G-attr from 100 images achieves over tenfold. See the Supplementary Material for details, including results on \textbf{AwA} and \textbf{CUB}.

\begin{table}[t]
\centering
{\small
\caption{Comparison of performances measured in AUSUC between GZSL with \textsc{word2vec} and GZSL with \textbf{G-attr} on the full \textbf{ImageNet} with 21,000 unseen classes. Few-shot results are averaged over 20 rounds.}
\label{detail_ImageNet_all}
\begin{tabular}{l|c|c|c|c} \hline
\multicolumn{1}{c|}{Method} & \multicolumn{4}{|c}{Flat hit@K} \\ \cline{2-5}
        & 1 & 5 & 10 & 20\\ \hline
\multicolumn{1}{c|}{\textsc{word2vec}} & 0.006 & 0.034 & 0.059 & 0.096 \\ \hline
G-attr from 1 image & 0.018\scriptsize{$\pm$0.0002}& 0.071\scriptsize{$\pm$0.0007}& 0.106\scriptsize{$\pm$0.0009} &0.150\scriptsize{$\pm$0.0011}\\
G-attr from 10 images & 0.050\scriptsize{$\pm$0.0002}& 0.184\scriptsize{$\pm$0.0003}& 0.263\scriptsize{$\pm$0.0004} &0.352\scriptsize{$\pm$0.0005}\\
G-attr from 100 images & 0.065\scriptsize{$\pm$0.0001}& 0.230\scriptsize{$\pm$0.0002}& 0.322\scriptsize{$\pm$0.0002} &0.421\scriptsize{$\pm$0.0002}\\
G-attr from all images & 0.067 & 0.236 & 0.329 & 0.429 \\  \hline
\end{tabular}
}
\end{table}


\section{Discussion}
\label{sDiscuss}
We investigate the problem of generalized zero-shot learning (GZSL). GZSL relaxes the unrealistic assumption in conventional zero-shot learning (ZSL) that test data belong only to unseen novel classes. In GZSL, test data might also come from seen classes and the labeling space is the union of both types of classes.  We show empirically that a straightforward application of classifiers provided by existing ZSL approaches does not perform well in the setting of GZSL. Motivated by this, we propose a surprisingly simple but effective method to adapt ZSL approaches for GZSL. The main idea is to introduce a calibration factor to calibrate the classifiers for both seen and unseen classes so as to balance two conflicting forces: recognizing  data from seen classes and those from unseen ones. We develop a new performance metric called the Area Under Seen-Unseen accuracy Curve to characterize this trade-off. We demonstrate the utility of this metric by analyzing existing ZSL approaches applied to the generalized setting. Extensive empirical studies  reveal strengths and weaknesses of those approaches on three well-studied benchmark datasets, including the large-scale ImageNet with more than 20,000 unseen categories. We complement our comparative studies in learning methods by further establishing an upper bound on the performance limit of GZSL. In particular, our idea is to use class-representative visual features as the idealized semantic embeddings. We show that there is a large gap between the performance of existing approaches and the performance limit, suggesting that improving the quality of class semantic embeddings is vital to improving ZSL.\\[1em]
\textbf{Acknowledgements} {\small B.G. is partially supported by NSF IIS-1566511. Others are partially supported by USC Graduate Fellowship, NSF IIS-1065243, 1451412, 1513966, 1208500, CCF-1139148, a Google Research Award, an Alfred. P. Sloan Research Fellowship and ARO\# W911NF-12-1-0241 and W911NF-15-1-0484.}

{\small
\bibliographystyle{splncs}
\bibliography{main}

\begin{thebibliography}{10}

\bibitem{SudderthJ08}
Sudderth, E.B., Jordan, M.I.:
\newblock Shared segmentation of natural scenes using dependent pitman-yor
  processes.
\newblock In: NIPS. (2008)

\bibitem{SalakhutdinovTT11}
Salakhutdinov, R., Torralba, A., Tenenbaum, J.:
\newblock Learning to share visual appearance for multiclass object detection.
\newblock In: CVPR. (2011)

\bibitem{zhu2014capturing}
Zhu, X., Anguelov, D., Ramanan, D.:
\newblock Capturing long-tail distributions of object subcategories.
\newblock In: CVPR. (2014)

\bibitem{LampertNH09}
Lampert, C.H., Nickisch, H., Harmeling, S.:
\newblock Learning to detect unseen object classes by between-class attribute
  transfer.
\newblock In: CVPR. (2009)

\bibitem{FarhadiEHF09}
Farhadi, A., Endres, I., Hoiem, D., Forsyth, D.:
\newblock Describing objects by their attributes.
\newblock In: CVPR. (2009)

\bibitem{ParikhG11}
Parikh, D., Grauman, K.:
\newblock Relative attributes.
\newblock In: ICCV. (2011)

\bibitem{MikolovCCD13}
Mikolov, T., Chen, K., Corrado, G.S., Dean, J.:
\newblock Efficient estimation of word representations in vector space.
\newblock In: ICLR Workshops. (2013)

\bibitem{FromeCSBDRM13}
Frome, A., Corrado, G.S., Shlens, J., Bengio, S., Dean, J., Ranzato, M.,
  Mikolov, T.:
\newblock Devise: A deep visual-semantic embedding model.
\newblock In: NIPS. (2013)

\bibitem{SocherGMN13}
Socher, R., Ganjoo, M., Manning, C.D., Ng, A.:
\newblock Zero-shot learning through cross-modal transfer.
\newblock In: NIPS. (2013)

\bibitem{PalatucciPHM09}
Palatucci, M., Pomerleau, D., Hinton, G.E., Mitchell, T.M.:
\newblock Zero-shot learning with semantic output codes.
\newblock In: NIPS. (2009)

\bibitem{YuA10}
Yu, X., Aloimonos, Y.:
\newblock Attribute-based transfer learning for object categorization with
  zero/one training example.
\newblock In: ECCV. (2010)

\bibitem{RohrbachSS11}
Rohrbach, M., Stark, M., Schiele, B.:
\newblock Evaluating knowledge transfer and zero-shot learning in a large-scale
  setting.
\newblock In: CVPR. (2011)

\bibitem{KankuekulKTH12}
Kankuekul, P., Kawewong, A., Tangruamsub, S., Hasegawa, O.:
\newblock Online incremental attribute-based zero-shot learning.
\newblock In: CVPR. (2012)

\bibitem{AkataPHS13}
Akata, Z., Perronnin, F., Harchaoui, Z., Schmid, C.:
\newblock Label-embedding for attribute-based classification.
\newblock In: CVPR. (2013)

\bibitem{YuCFSC13}
Yu, F.X., Cao, L., Feris, R.S., Smith, J.R., Chang, S.F.:
\newblock Designing category-level attributes for discriminative visual
  recognition.
\newblock In: CVPR. (2013)

\bibitem{MensinkGS14}
Mensink, T., Gavves, E., Snoek, C.G.:
\newblock Costa: Co-occurrence statistics for zero-shot classification.
\newblock In: CVPR. (2014)

\bibitem{NorouziMBSSFCD14}
Norouzi, M., Mikolov, T., Bengio, S., Singer, Y., Shlens, J., Frome, A.,
  Corrado, G.S., Dean, J.:
\newblock Zero-shot learning by convex combination of semantic embeddings.
\newblock In: ICLR. (2014)

\bibitem{JayaramanG14}
Jayaraman, D., Grauman, K.:
\newblock Zero-shot recognition with unreliable attributes.
\newblock In: NIPS. (2014)

\bibitem{AlHalahS15}
Al-Halah, Z., Stiefelhagen, R.:
\newblock How to transfer? zero-shot object recognition via hierarchical
  transfer of semantic attributes.
\newblock In: WACV. (2015)

\bibitem{AkataRWLS15}
Akata, Z., Reed, S., Walter, D., Lee, H., Schiele, B.:
\newblock Evaluation of output embeddings for fine-grained image
  classification.
\newblock In: CVPR. (2015)

\bibitem{FuHXG15}
Fu, Y., Hospedales, T.M., Xiang, T., Gong, S.:
\newblock Transductive multi-view zero-shot learning.
\newblock TPAMI (2015)

\bibitem{FuXKG15}
Fu, Z., Xiang, T., Kodirov, E., Gong, S.:
\newblock Zero-shot object recognition by semantic manifold distance.
\newblock In: CVPR. (2015)

\bibitem{LiGS15}
Li, X., Guo, Y., Schuurmans, D.:
\newblock Semi-supervised zero-shot classification with label representation
  learning.
\newblock In: ICCV. (2015)

\bibitem{Bernardino15}
Romera-Paredes, B., Torr, P.H.S.:
\newblock An embarrassingly simple approach to zero-shot learning.
\newblock In: ICML. (2015)

\bibitem{KodirovXFG15}
Kodirov, E., Xiang, T., Fu, Z., Gong, S.:
\newblock Unsupervised domain adaptation for zero-shot learning.
\newblock In: ICCV. (2015)

\bibitem{ZhangS15}
Zhang, Z., Saligrama, V.:
\newblock Zero-shot learning via semantic similarity embedding.
\newblock In: ICCV. (2015)

\bibitem{zhang2016classifying}
Zhang, Z., Saligrama, V.:
\newblock Zero-shot learning via joint latent similarity embedding.
\newblock In: CVPR. (2016)

\bibitem{ChangpinyoCGS16}
Changpinyo, S., Chao, W.L., Gong, B., Sha, F.:
\newblock Synthesized classifiers for zero-shot learning.
\newblock In: CVPR. (2016)

\bibitem{deng2009imagenet}
Deng, J., Dong, W., Socher, R., Li, L.J., Li, K., Fei-Fei, L.:
\newblock Imagenet: A large-scale hierarchical image database.
\newblock In: CVPR. (2009)

\bibitem{MensinkVPC12}
Mensink, T., Verbeek, J., Perronnin, F., Csurka, G.:
\newblock Metric learning for large scale image classification: Generalizing to
  new classes at near-zero cost.
\newblock In: ECCV. (2012)

\bibitem{Tang2010optimizing}
Tang, K.D., Tappen, M.F., Sukthankar, R., Lampert, C.H.:
\newblock Optimizing one-shot recognition with micro-set learning.
\newblock In: CVPR. (2010)

\bibitem{gan2016recognizing}
Gan, C., Yang, Y., Zhu, L., Zhao, D., Zhuang, Y.:
\newblock Recognizing an action using its name: A knowledge-based approach.
\newblock IJCV (2016)  1--17

\bibitem{ElhoseinySE13}
Elhoseiny, M., Saleh, B., Elgammal, A.:
\newblock Write a classifier: Zero-shot learning using purely textual
  descriptions.
\newblock In: ICCV. (2013)

\bibitem{LeiSFS15}
Lei~Ba, J., Swersky, K., Fidler, S., Salakhutdinov, R.:
\newblock Predicting deep zero-shot convolutional neural networks using textual
  descriptions.
\newblock In: ICCV. (2015)

\bibitem{scheirer2013toward}
Scheirer, W.J., de~Rezende~Rocha, A., Sapkota, A., Boult, T.E.:
\newblock Toward open set recognition.
\newblock TPAMI \textbf{35}(7) (2013)  1757--1772

\bibitem{scheirer2014probability}
Scheirer, W.J., Jain, L.P., Boult, T.E.:
\newblock Probability models for open set recognition.
\newblock TPAMI \textbf{36}(11) (2014)  2317--2324

\bibitem{jain2014multi}
Jain, L.P., Scheirer, W.J., Boult, T.E.:
\newblock Multi-class open set recognition using probability of inclusion.
\newblock In: ECCV. (2014)  393--409

\bibitem{LampertNH14}
Lampert, C.H., Nickisch, H., Harmeling, S.:
\newblock Attribute-based classification for zero-shot visual object
  categorization.
\newblock TPAMI \textbf{36}(3) (2014)  453--465

\bibitem{WahCUB_200_2011}
Wah, C., Branson, S., Welinder, P., Perona, P., Belongie, S.:
\newblock {The Caltech-UCSD Birds-200-2011 Dataset}.
\newblock Technical Report CNS-TR-2011-001, California Institute of Technology
  (2011)

\bibitem{KriegelKSZ09}
Kriegel, H.P., Kr{\"o}ger, P., Schubert, E., Zimek, A.:
\newblock {LoOP: local outlier probabilities}.
\newblock In: CIKM. (2009)

\bibitem{ILSVRC15}
Russakovsky, O., Deng, J., Su, H., Krause, J., Satheesh, S., Ma, S., Huang, Z.,
  Karpathy, A., Khosla, A., Bernstein, M., Berg, A.C., Fei-Fei, L.:
\newblock {ImageNet Large Scale Visual Recognition Challenge}.
\newblock IJCV (2015)

\bibitem{MikolovSCCD13}
Mikolov, T., Sutskever, I., Chen, K., Corrado, G.S., Dean, J.:
\newblock Distributed representations of words and phrases and their
  compositionality.
\newblock In: NIPS. (2013)

\bibitem{SzegedyLJSRAEVR14}
Szegedy, C., Liu, W., Jia, Y., Sermanet, P., Reed, S., Anguelov, D., Erhan, D.,
  Vanhoucke, V., Rabinovich, A.:
\newblock Going deeper with convolutions.
\newblock In: CVPR. (2015)

\bibitem{jia2014caffe}
Jia, Y., Shelhamer, E., Donahue, J., Karayev, S., Long, J., Girshick, R.,
  Guadarrama, S., Darrell, T.:
\newblock Caffe: Convolutional architecture for fast feature embedding.
\newblock In: ACM Multimedia. (2014)

\bibitem{CrammerS02}
Crammer, K., Singer, Y.:
\newblock On the algorithmic implementation of multiclass kernel-based vector
  machines.
\newblock JMLR \textbf{2} (2002)  265--292

\bibitem{PattersonH14}
Patterson, G., Xu, C., Su, H., Hays, J.:
\newblock The sun attribute database: Beyond categories for deeper scene
  understanding.
\newblock IJCV \textbf{108}(1-2) (2014)  59--81

\end{thebibliography}
}

\appendix
\title{Supplementary Material: An Empirical Study and Analysis of Generalized Zero-Shot Learning for Object Recognition in the Wild}
\titlerunning{An Empirical Study \& Analysis of GZSL for Object Recognition in the Wild}
\authorrunning{Wei-Lun Chao$^*$, Soravit Changpinyo$^*$, Boqing Gong, and Fei Sha}

\author{Wei-Lun Chao$^*$$^1$, Soravit Changpinyo\thanks{\hspace{4pt}Equal contribution.}$^1$, Boqing Gong$^2$, and Fei Sha$^3$}
\institute{$^1$Dept. of Computer Science, U. of Southern California, United States\\
$^2$Center for Research in Computer Vision, U. of Central Florida, United States\\
$^3$Dept. of Computer Science, U. of California, Los Angeles, United States\\
\email{\{weilunc, schangpi\}@usc.edu, bgong@crcv.ucf.edu, feisha@cs.ucla.edu}}
\maketitle

$\newline$
This Supplementary Material provides the following additional details, results, and analysis (along with their corresponding sections of the main text).
\begin{itemize}
	\item Section~\ref{ssCV}: Hyper-parameter tuning strategies (Section 4.2 of the main text).
	\item Section~\ref{ssNovelty}: Novelty detection approaches: details and additional results (Section 4.3 and 5.2 of the main text).
	\item Section~\ref{ssExp}: Comparison between zero-shot learning approaches: additional ZSL algorithm, dataset, and results (Section 5 of the main text).
	\item Section~\ref{ssAnalysis}: Analysis on (generalized) zero-shot learning: details and additional results (Section 6 of the main text).
\end{itemize}

\section{Hyper-parameter tuning strategies}
\label{ssCV}

\subsection{Cross-validation with AUSUC}
\label{CVAUC}
In Section 4.2 of the main text, we introduce the Area Under Seen-Unseen accuracy Curve (AUSUC), which is analogous to many metrics in computer vision and machine learning that balance two conflicting (sub)metrics, such as area under ROC. To tune the hyper-parameters based on this metric\footnote{AUSUC is computed by varying the $\gamma$ factor within a range. If a single $\gamma$ is desired, another measure such as ``F-score'' balancing $A_{\mathcal{U} \rightarrow \mathcal{T}}$ and $A_{\mathcal{S} \rightarrow \mathcal{T}}$ can be used. One can also assume a prior probability of whether any instance is seen or unseen to select the factor.}, we simulate the generalized zero-shot learning setting during cross-validation. 

Concretely, we split the training data into 5 folds A1, A2, A3, A4 and A5 so that the \emph{class labels of these folds are disjoint}. We further split 80\% and 20\% of data from each fold (A1-A5, respectively) into \emph{pseudo-train} and \emph{pseudo-test} sets, respectively. We then combine the \emph{pseudo-train} sets of four folds (for example, A1-A4) for training, and validate on (i) the \emph{pseudo-test} sets of such four folds (i.e., A1-A4) and (ii) the \emph{pseudo-train} set of the remaining fold (i.e., A5). That is, the remaining fold serves as the \emph{pseudo-unseen} classes in cross-validation. We repeat this process for 5 rounds --- each round selects a fold as the ``remaining" fold, and computes AUSUC on the corresponding validation set. Finally, the average of AUSUCs over all rounds is used to select hyper-parameters.

\subsection{Comparison to an alternative strategy}
\label{CVacc}
Another strategy for hyper-parameter tuning is to find two sets of hyper-paramet-ers: one optimized for seen classes and the other for unseen classes.
The standard cross-validation technique, where $A_{\mathcal{S} \rightarrow \mathcal{S}}$ is optimized, can be used for the former.
For the latter, it has been shown that the class-wise cross-validation technique~\cite{ChangpinyoCGS16,ZhangS15,ElhoseinySE13}, where the conventional zero-shot learning task is simulated, outperforms the standard technique \cite{ChangpinyoCGS16}. In this case, $A_{\mathcal{U} \rightarrow \mathcal{U}}$ is optimized. We thus use the first set of hyper-parameters to construct the scoring functions for the seen classes, and use the second set for the unseen classes (cf. Section 3.2 and 3.3 of the main text).

In this subsection, we show that the strategy that jointly optimizes hyper-parameters based on AUSUC in most cases leads to better models for GZSL than the strategy that optimizes seen and unseen classifiers' performances separately. On \textbf{AwA} and \textbf{CUB}, we perform 5-fold cross-validation based on the two strategies and compare the performance of those selected models in Table~\ref{tCV}. In general, cross-validation based on AUSUC leads to better models for GZSL. The exceptions are ConSE on \textbf{AwA} and DAP on \textbf{CUB}. 

\begin{table}
\centering
{\small
\caption{Comparison of performance measured in AUSUC between two cross-validation strategies on \textbf{AwA} and \textbf{CUB}. One strategy is based on accuracies ($A_{\mathcal{S} \rightarrow \mathcal{S}}$ and $A_{\mathcal{U} \rightarrow \mathcal{U}}$) and the other is based on AUSUC. See text for details.}
\label{tCV}
\begin{tabular}{c|c|c||c|c} \hline
 & \multicolumn{2}{|c||}{\textbf{AwA}} & \multicolumn{2}{|c}{\textbf{CUB}} \\ \cline{2-5}
Method & \multicolumn{2}{|c||}{CV strategies} & \multicolumn{2}{|c}{CV strategies} \\ \cline{2-5}
 & Accuracies & AUSUC & Accuracies & AUSUC \\ 
\hline
DAP~\cite{LampertNH14} & 0.341 & \textbf{0.366} & \textbf{0.202} & 0.194 \\
IAP~\cite{LampertNH14} & 0.366 & \textbf{0.394} & 0.194 & \textbf{0.199}  \\
ConSE~\cite{NorouziMBSSFCD14} & \textbf{0.443} & 0.428 & 0.190 & \textbf{0.212} \\
{SynC$^\textrm{o-vs-o}$}~\cite{ChangpinyoCGS16} & 0.539 & \textbf{0.568} & 0.324 & \textbf{0.336}\\
{SynC$^\textrm{struct}$}~\cite{ChangpinyoCGS16} & 0.551 & \textbf{0.583} & 0.356 & 0.356 \\ \hline
\end{tabular}
}
\end{table}


\section{Novelty detection approaches: details and additional results }
\label{ssNovelty}

In Section 4.3, we describe alternative approaches to calibrated stacking that are based on the idea of novelty detection. 
The two novelty detection approaches --- Gaussian and LoOP \cite{KriegelKSZ09} --- have been explored by \cite{SocherGMN13}.
In this section, we provide details and additional results on these approaches.

\subsection{Algorithms}

In \cite{SocherGMN13}, Socher et al. first learn a mapping from the visual feature space to the semantic embedding space. The novelty detection is then performed in this semantic space. Below we describe how to compute novelty scores under Gaussian and LoOP models.

\subsubsection{Gaussian}
Training examples of seen classes are first mapped into the semantic space and modeled by a Gaussian mixture model --- each class is parameterized by a mean vector and an isometric covariance matrix. The mean vector is set to be the class' semantic embedding and the covariance matrix is set to be the covariance of all mapped training examples of that class.
The novelty score of a test data point is then its negative log probability value under this mixture model.

\subsubsection{LoOP}
Let $X_\sS$ be the set of all the mapped training examples from seen classes.
For a test sample $\vx$ (also mapped into the semantic space), a context set $C(\vx) \subseteq X_\sS$ of $k$ nearest neighbors is first defined.
The probabilistic set distance $pdist$ from $\vx$ to all the points in $C(\vx)$ is then computed as follows
\begin{align}
pdist_{\lambda}(\vx, C(\vx)) = \lambda \sqrt{ \frac{\sum_{\vx' \in C(x)} d(\vx,\vx')^2}{|C(\vx)|} },
\end{align}
where $d(\vx,\vx')$ is chosen to be the Euclidean distance function.
Such a distance is then used to define the local outlier factor
\begin{align}
lof_{\lambda}(\vx) = \frac{ pdist_{\lambda}(\vx, C(\vx)) }{ \expect{\vx' \in C(\vx)}{pdist_{\lambda}(\vx', C(\vx'))} } - 1.
\end{align}
Finally, the Local Outlier Probability (LoOP), which can be viewed as the novelty score, is computed as
\begin{align}
LoOP(\vx) = \max \left\{ 0, \textrm{erf}\left(\frac{lof_{\lambda}(\vx)}{Z_{\lambda}(X_\sS)}\right) \right\},
\end{align}
where erf is the Gauss error function and $Z_{\lambda}(X_\sS) = \lambda \sqrt{\expect{\vx' \in X_\sS}{(lof_{\lambda}(\vx'))^2}}$ is the normalization constant.

\subsection{Implementation details}
We use the code provided by Socher et al. \cite{SocherGMN13} and follow their settings.
In particular, we train a two-layer neural network with the same loss function as in \cite{SocherGMN13} to learn a mapping from the visual feature space to the semantic embedding space.
We tune the hyper-parameter $\lambda$ (a multiplier on the standard deviation) in LoOP jointly with other hyper-parameters of zero-shot learning approaches --- although we empirically observe that $\lambda$ does not significantly affect the novelty detection rankings, consistent with the observations made by \cite{KriegelKSZ09}.
Following \cite{SocherGMN13}, we set the number of neighbors (from the seen classes' examples) $k$ in LoOP to be 20.

\subsection{Additional results}
\begin{figure}
\centering
\includegraphics[width=.19\textwidth]{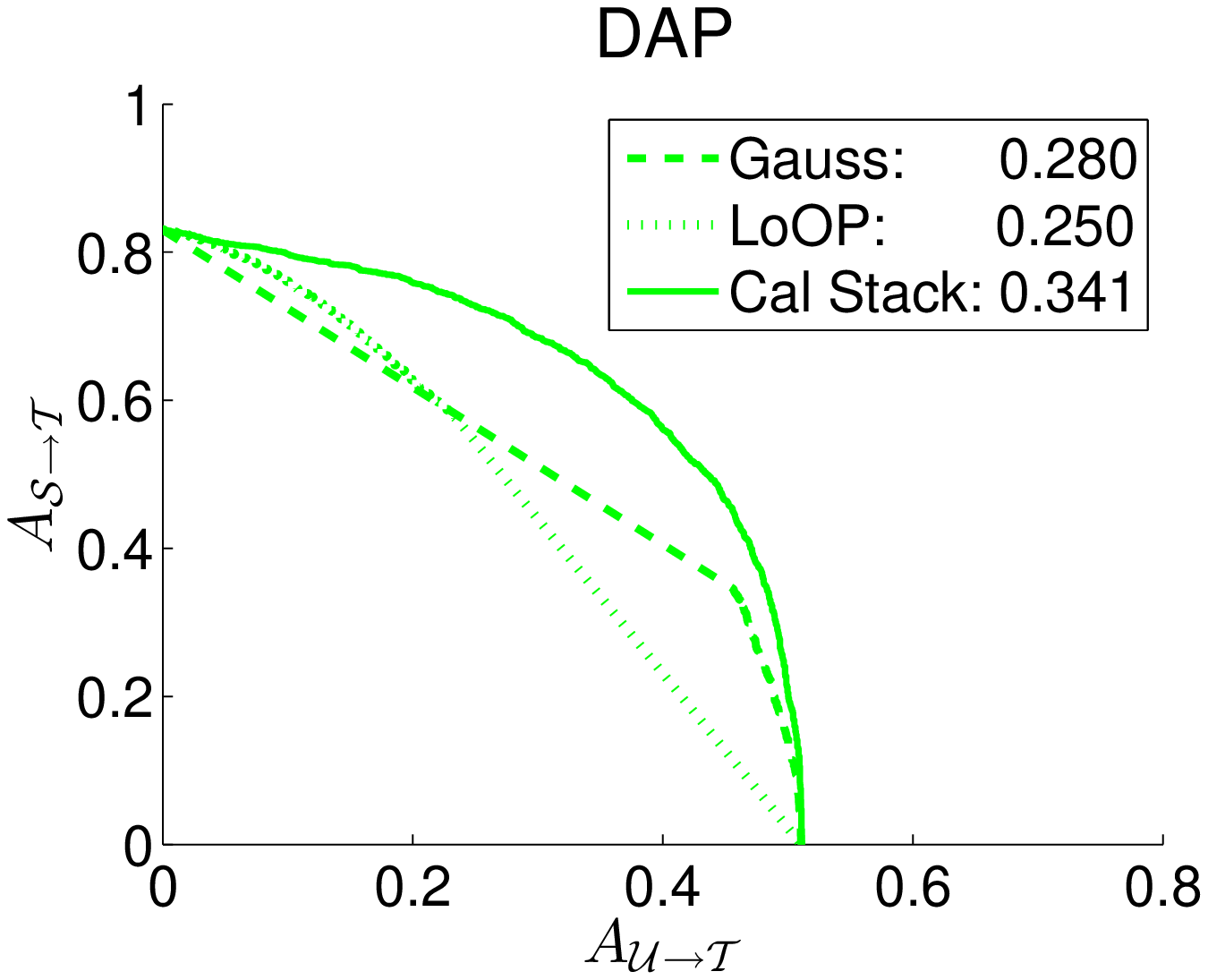}
\includegraphics[width=.19\textwidth]{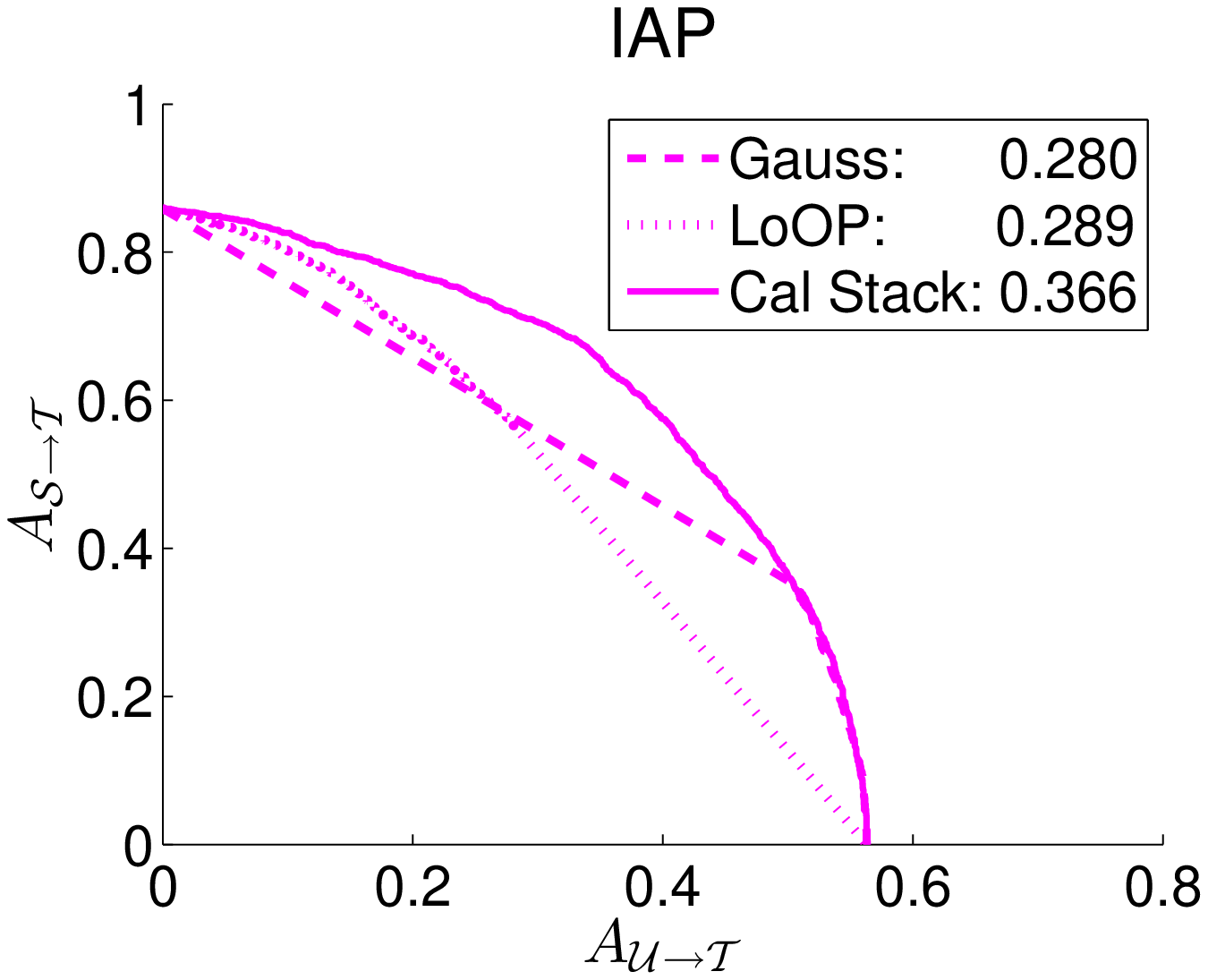}
\includegraphics[width=.19\textwidth]{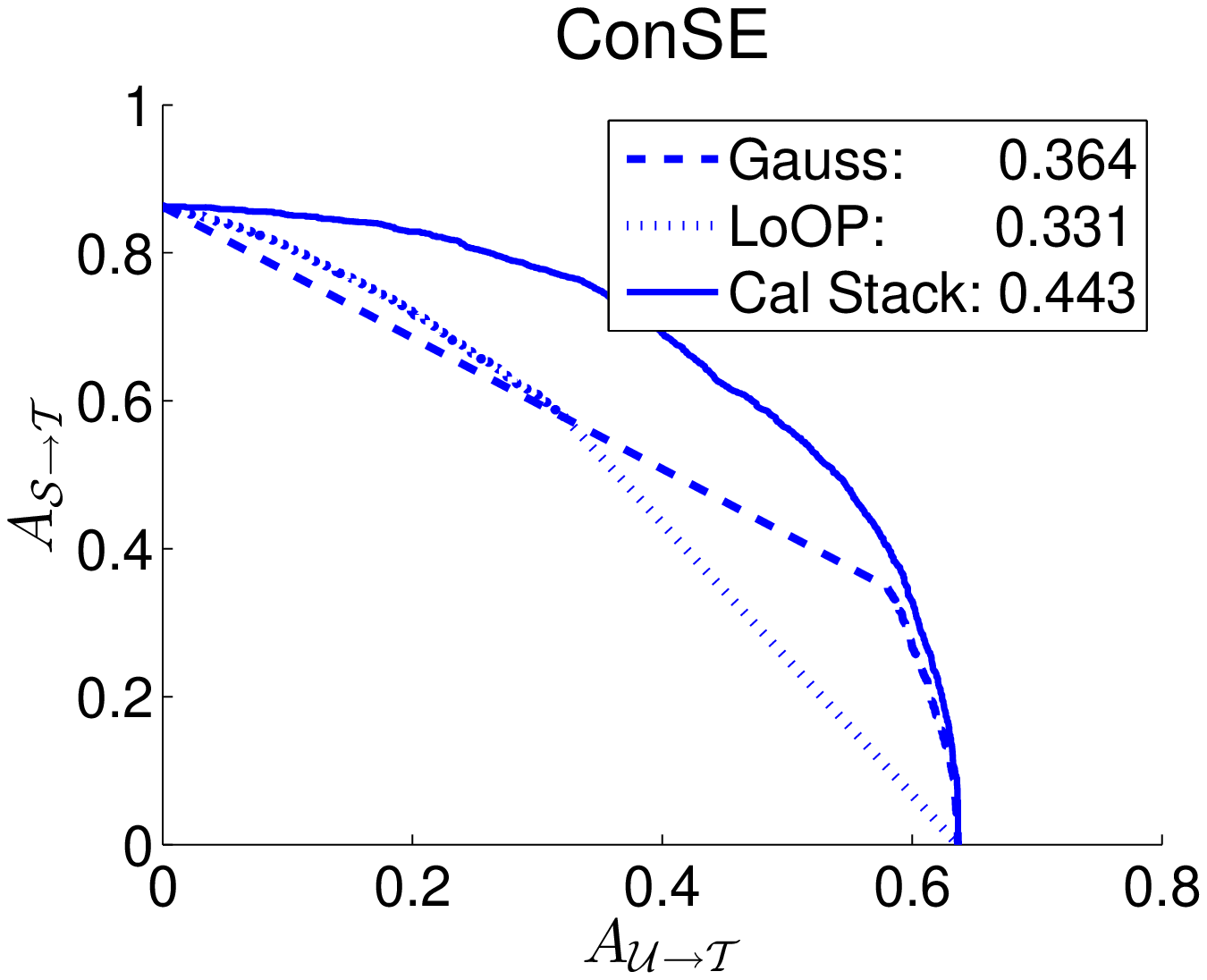}
\includegraphics[width=.19\textwidth]{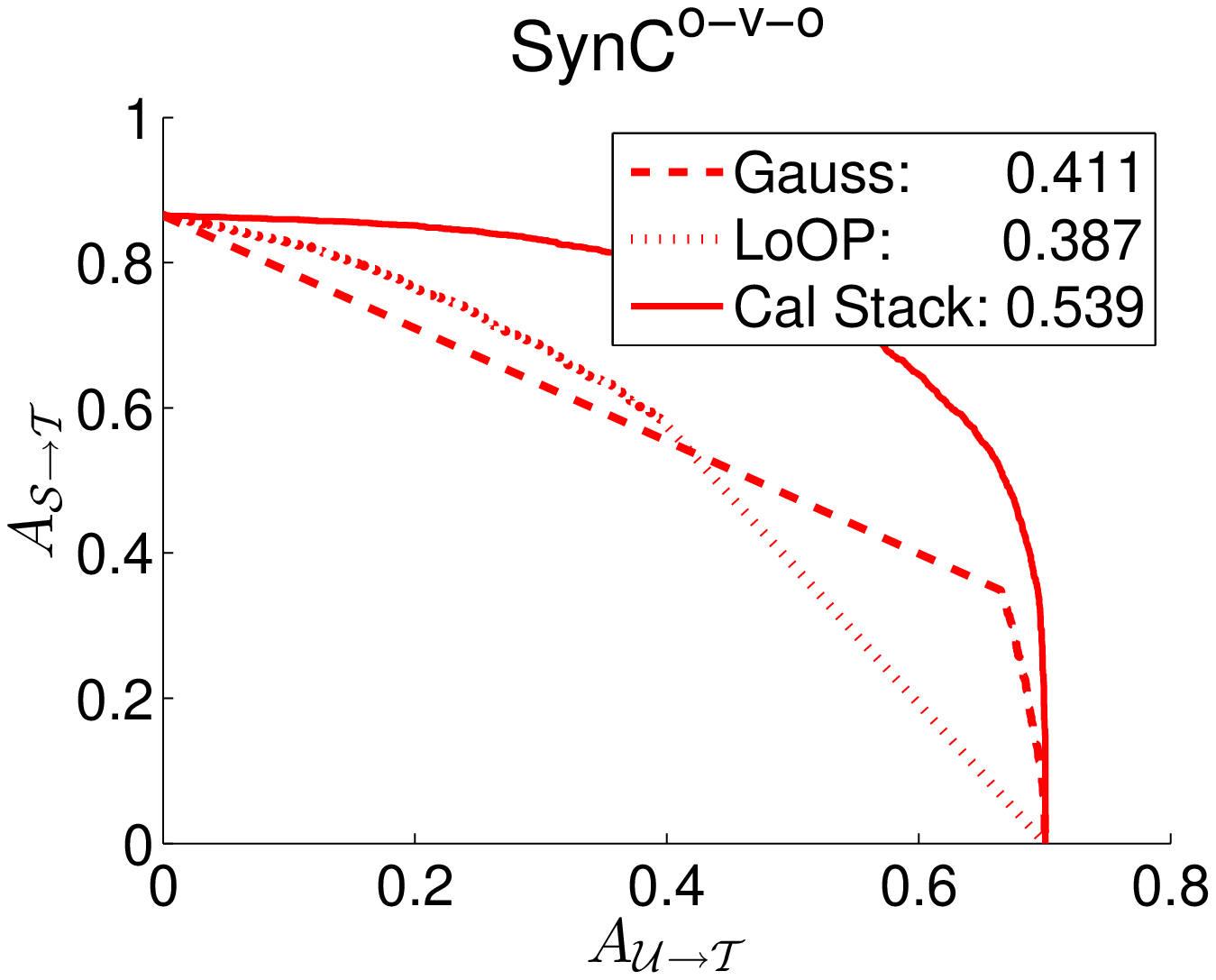}
\includegraphics[width=.19\textwidth]{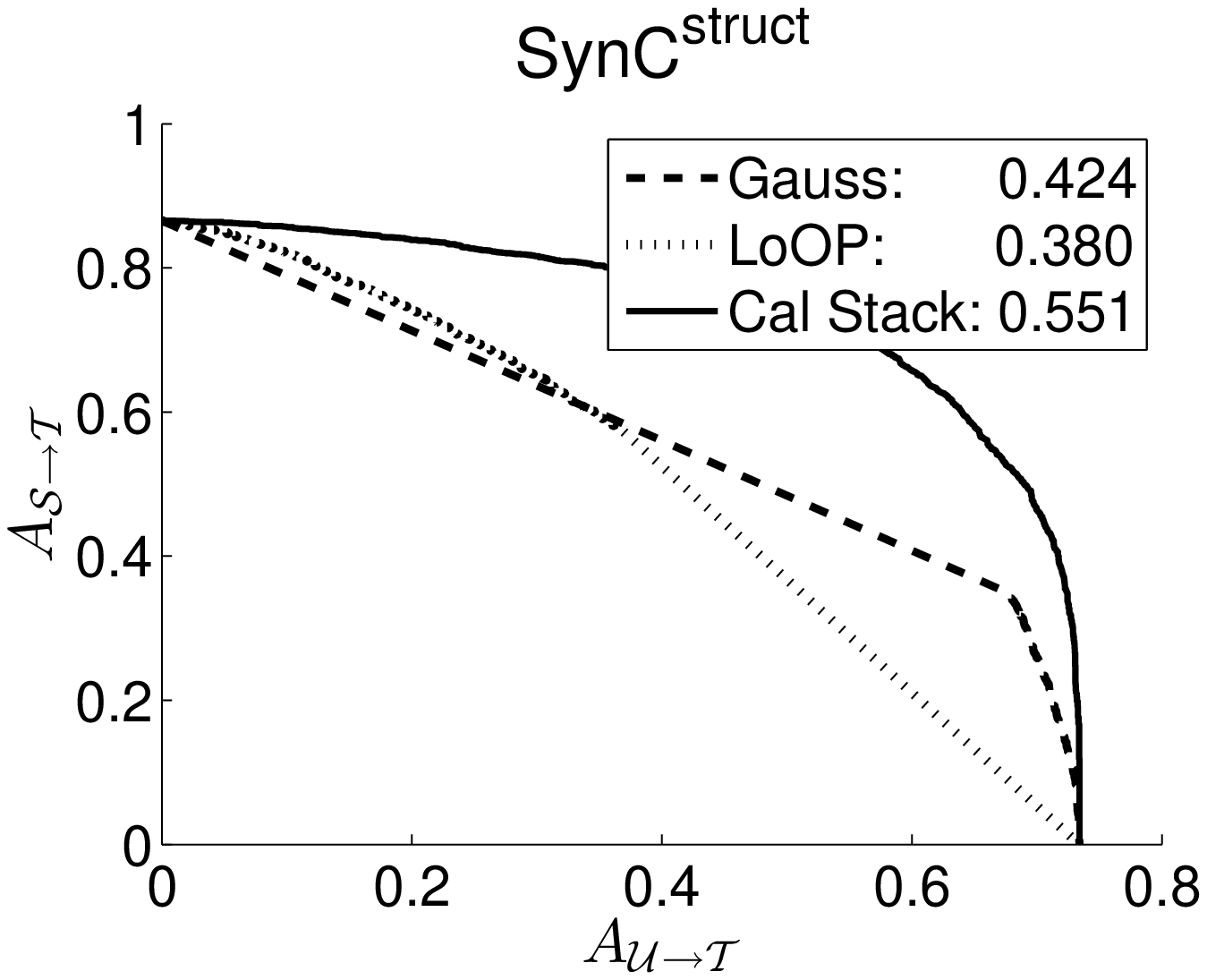}
\includegraphics[width=.19\textwidth]{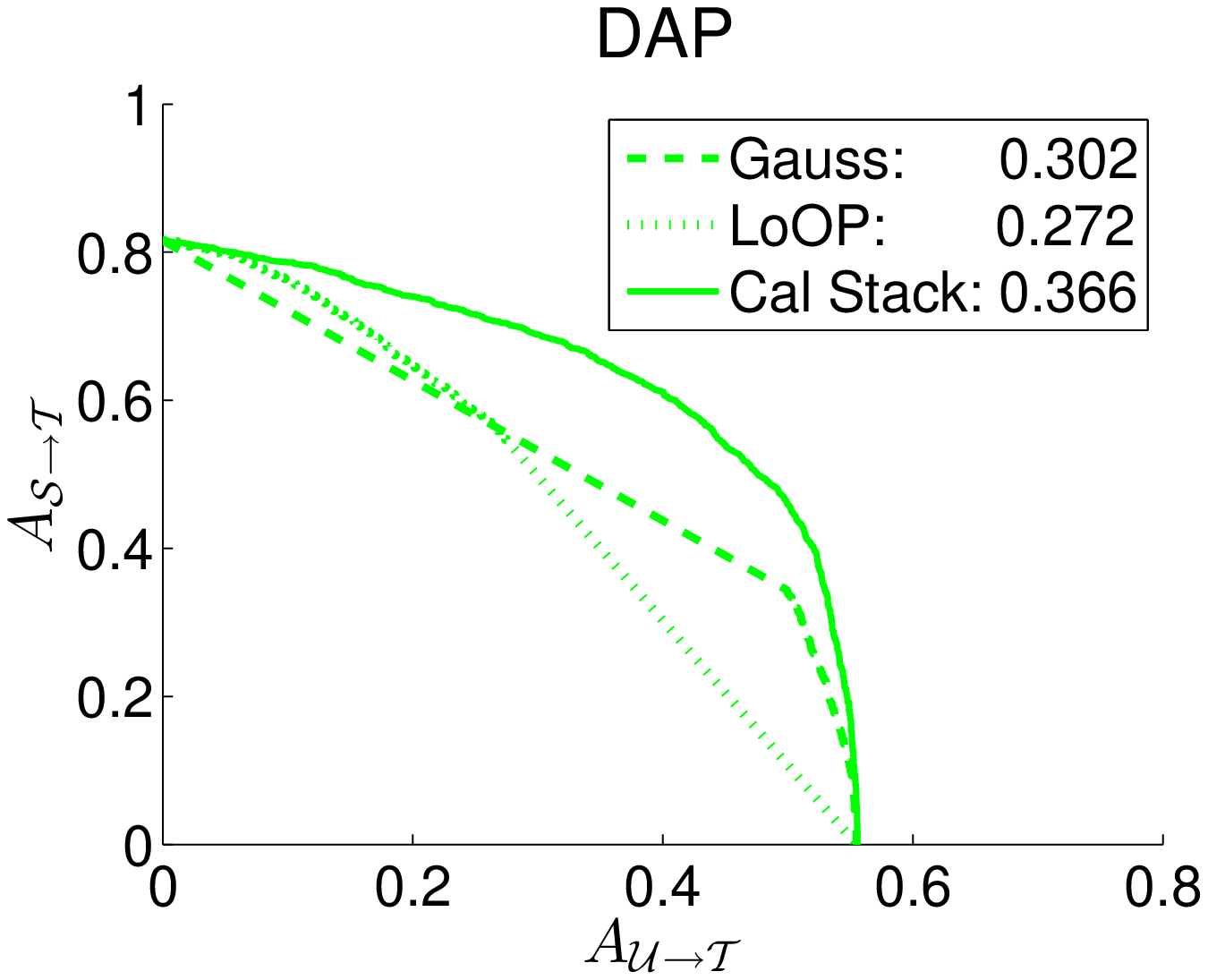}
\includegraphics[width=.19\textwidth]{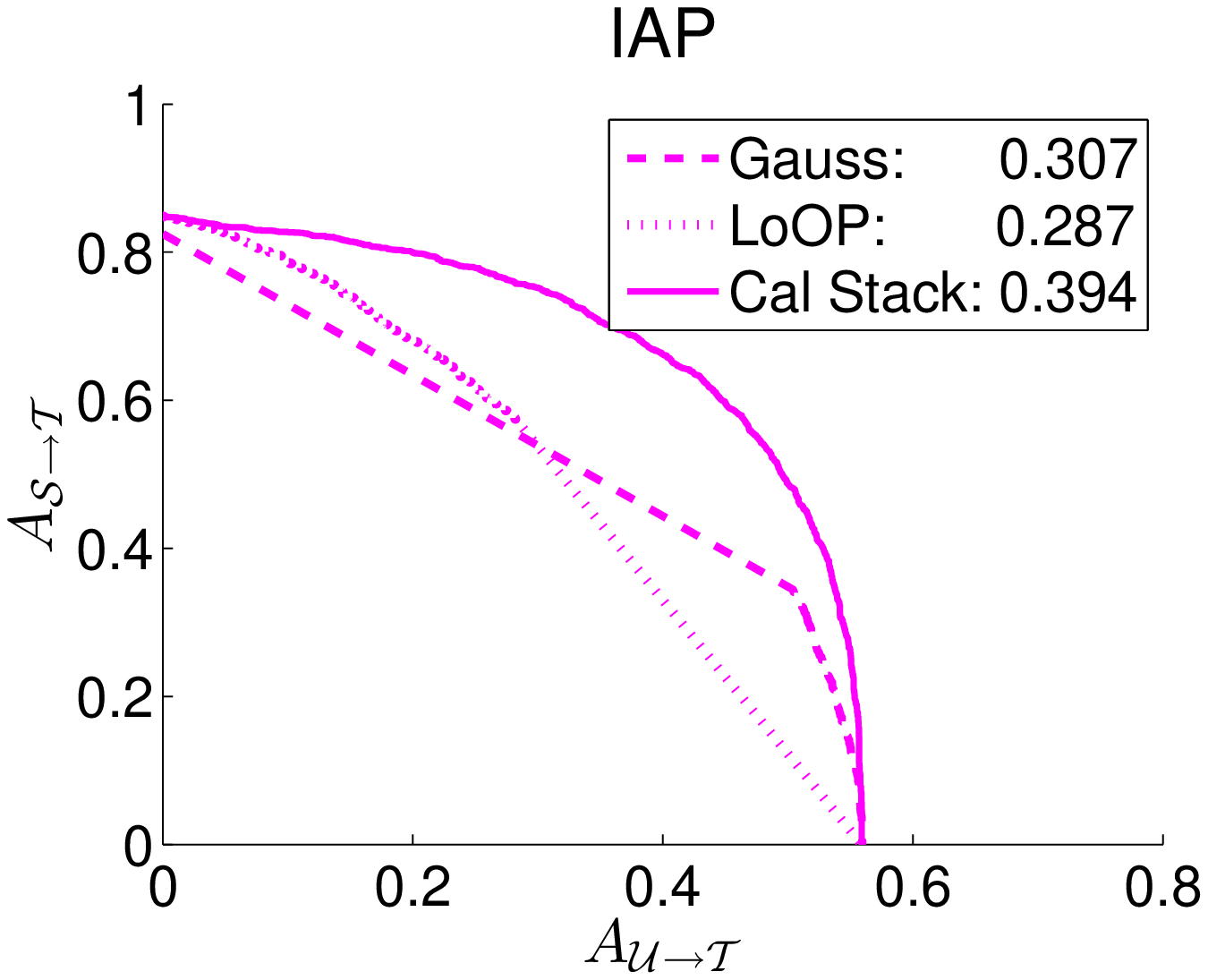}
\includegraphics[width=.19\textwidth]{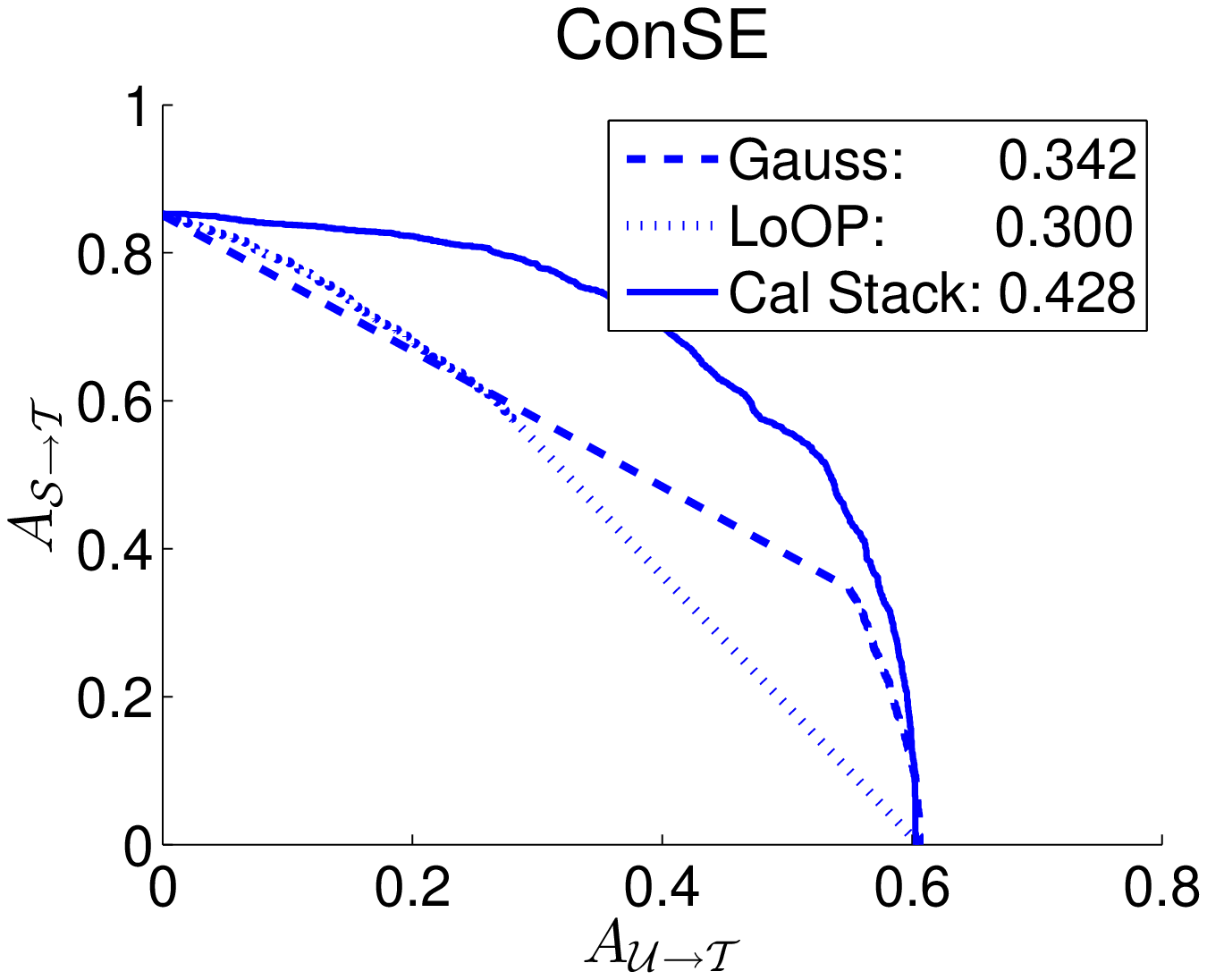}
\includegraphics[width=.19\textwidth]{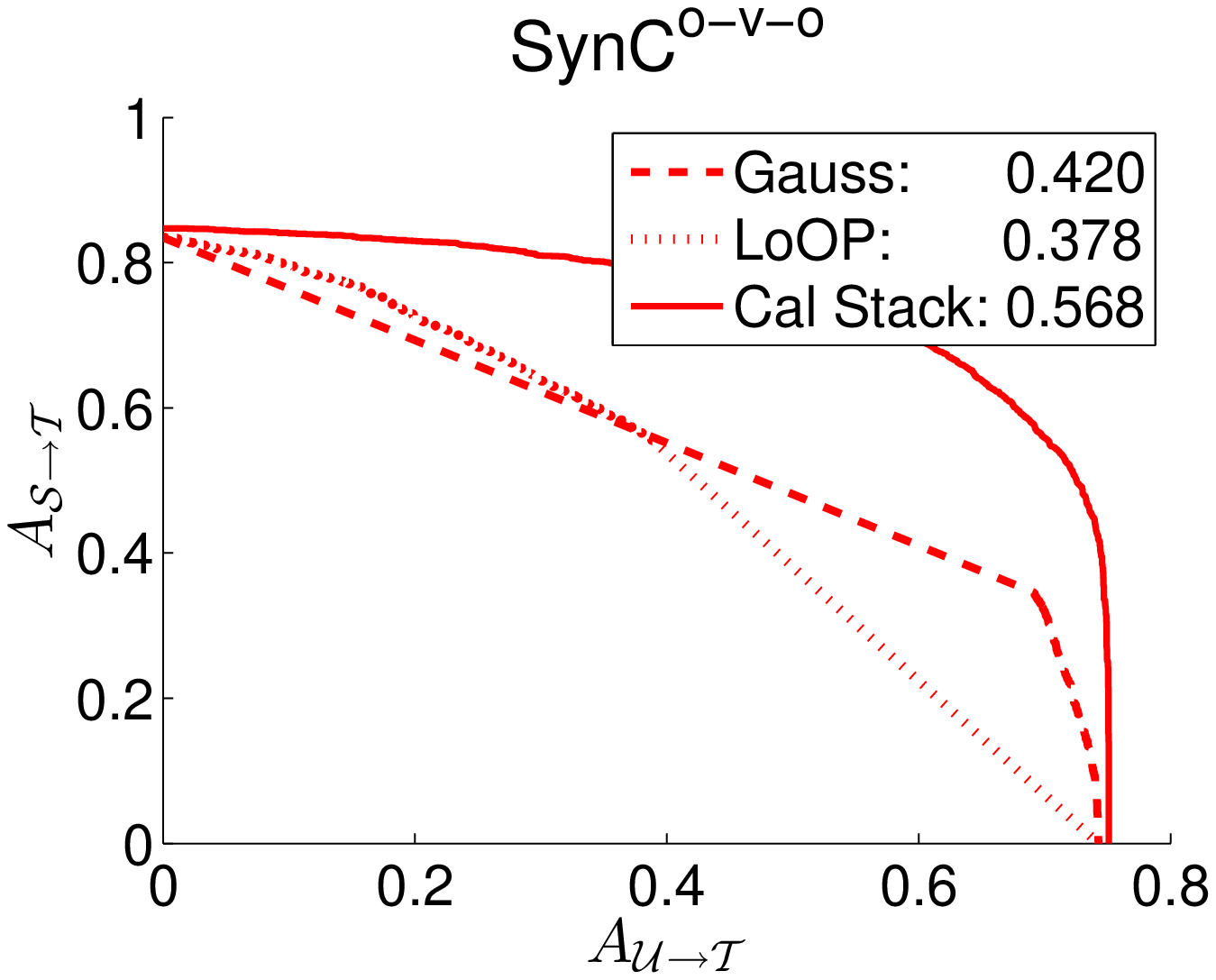}
\includegraphics[width=.19\textwidth]{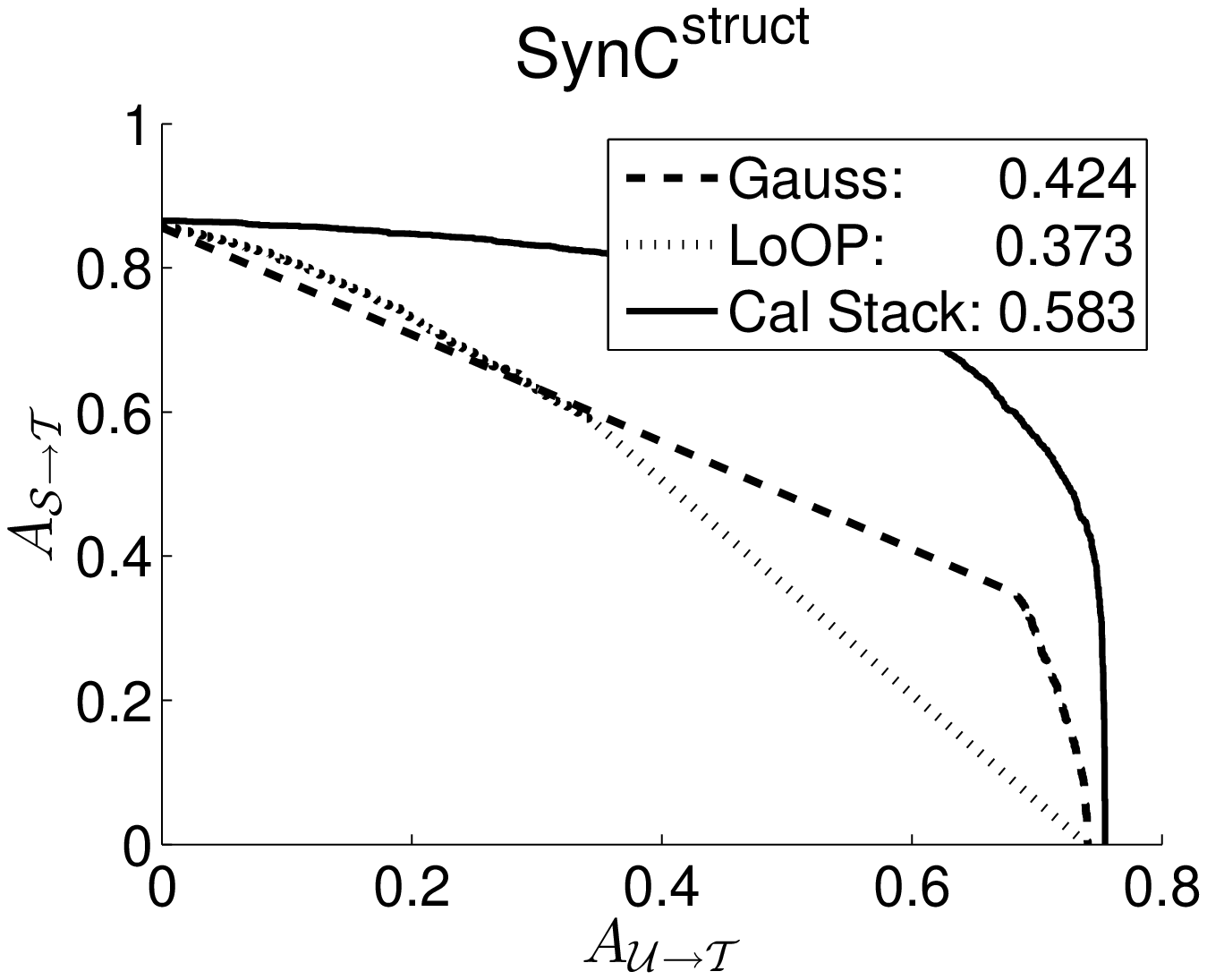}
\caption{\small Seen-Unseen accuracy Curves (SUC) for Gaussian (Gauss), LoOP, and calibrated stacking (Cal Stack) for all zero-shot learning approaches on \textbf{AwA}.
Hyper-parameters are cross-validated based on accuracies (top) and AUSUC (bottom). Calibrated stacking outperforms both Gaussian and LoOP in all cases.}
\label{fNov}
\end{figure}

In Section 5.2 and Table 3 of the main text, we compare Gaussian and LoOP to calibrated stacking, with hyper-parameters cross-validated to maximize AUSUC.
In Table~\ref{tNov_acc}, we show that calibrated stacking outperforms Gaussian and LoOP as well when hyper-parameters are cross-validated to maximize accuracies (cf. Section~\ref{CVacc}). We further show the SUCs of Gaussian, LoOP, and calibrated stacking on \textbf{AwA} in Fig.~\ref{fNov}. We observe the superior performance of calibrated stacking over Gaussian and LoOP across all zero-shot learning approaches, regardless of cross-validation strategies. Moreover, interestingly, we see that the curves for Gaussian and LoOP cross each other in such a way that implies that Gaussian has a tendency to classifying more data into ``unseen" categories (consistent with the observations reported by \cite{SocherGMN13}).

\begin{table}
\centering
{\small
\caption{Performance measured in AUSUC for novelty detection (Gaussian and LoOP) and calibrated stacking on \textbf{AwA} and \textbf{CUB}. Hyper-parameters are cross-validated to maximize accuracies. Calibrated stacking outperforms Gaussian and LoOP in all cases. Also, see Table 3 of the main text for the performance when hyper-parameters are cross-validated to directly maximize AUSUC.}
\label{tNov_acc}
\begin{tabular}{c|c|c|c||c|c|c} \hline
& \multicolumn{3}{|c||}{\textbf{AwA}} & \multicolumn{3}{|c}{\textbf{CUB}} \\ \cline{2-7}
Method  & \multicolumn{2}{|c|}{Novelty detection \cite{SocherGMN13}} & Calibrated  &  \multicolumn{2}{|c|}{Novelty detection \cite{SocherGMN13}} & Calibrated \\ \cline{2-3} \cline{5-6}
 & Gaussian & LoOP & Stacking & Gaussian & LoOP & Stacking \\  
\hline
DAP & 0.280 & 0.250 & \textbf{0.341} & 0.126 & 0.142 & \textbf{0.202}\\
IAP & 0.319 & 0.289 & \textbf{0.366} & 0.132 & 0.149 & \textbf{0.194}\\
ConSE & 0.364 & 0.331 & \textbf{0.443} & 0.131 & 0.141 & \textbf{0.190}\\
{SynC$^\textrm{o-vs-o}$} & 0.411 & 0.387 & \textbf{0.539} & 0.195 & 0.219 & \textbf{0.324} \\
{SynC$^\textrm{struct}$} & 0.424 & 0.380 & \textbf{0.551} & 0.199 & 0.225 & \textbf{0.356}\\ \hline
\end{tabular}
}
\end{table}

\section{Comparison between zero-shot learning approaches: additional algorithm, dataset, and results}
\label{ssExp}

\subsection{Additional results for Section 5.3 of the main text}
We provide additional SUC plots for comparing different zero-shot learning approaches, which complements Fig. 2 and Fig. 3 of Section 5.3 of the main text.

\begin{figure}
\centering
\includegraphics[width=.42\textwidth]{plots/AUSUC_CUB_1}
\includegraphics[width=.42\textwidth]{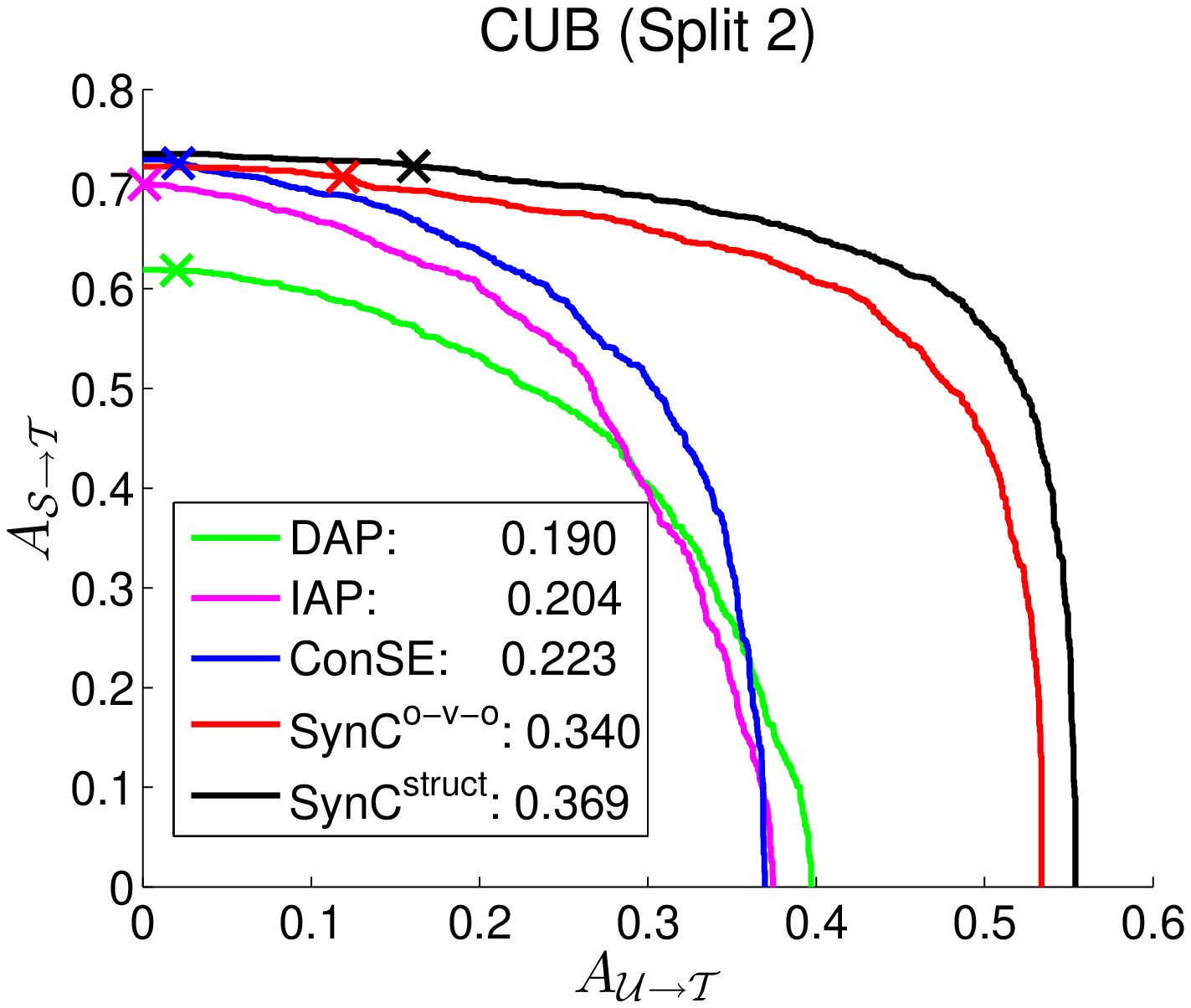}
\includegraphics[width=.42\textwidth]{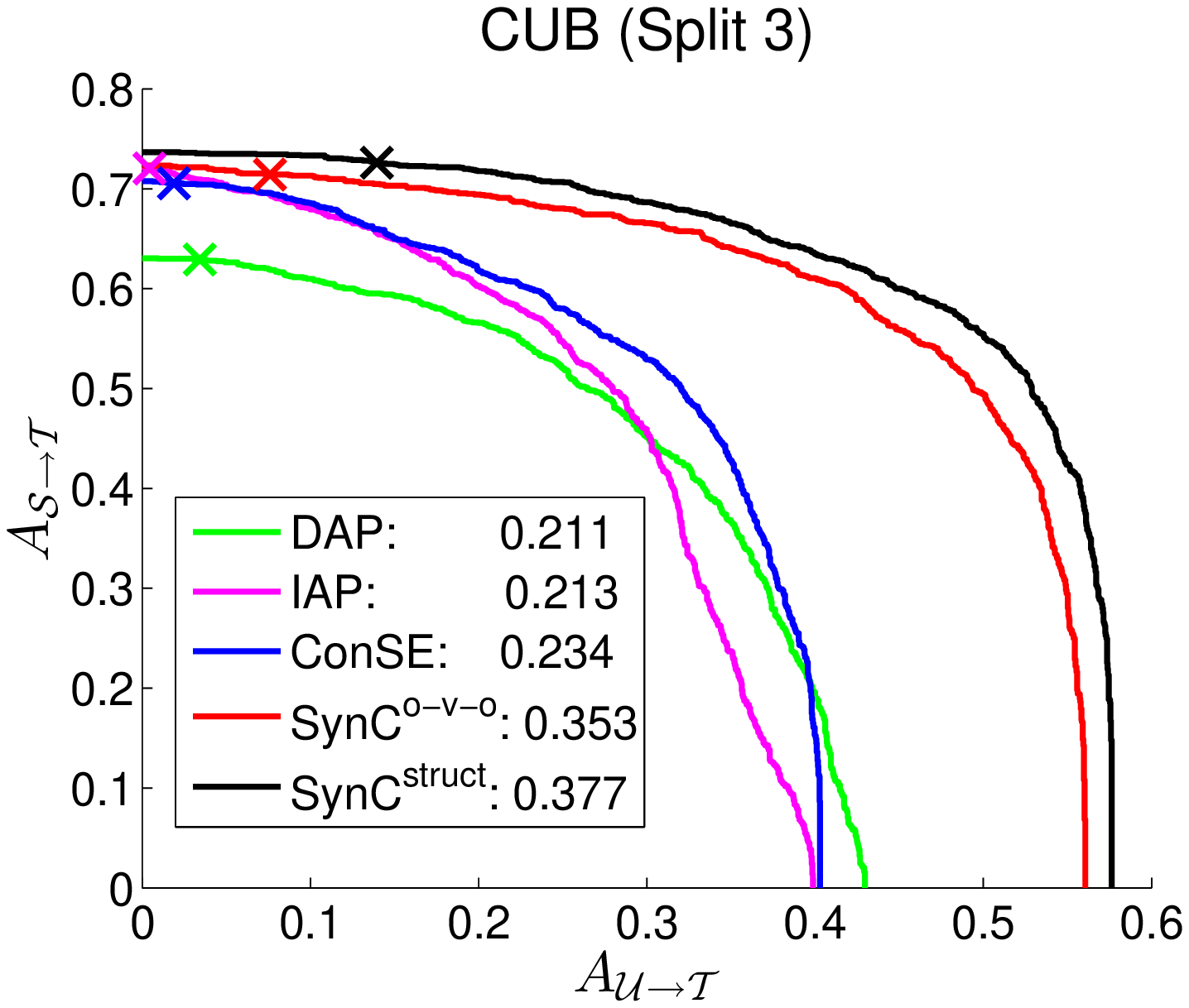}
\includegraphics[width=.42\textwidth]{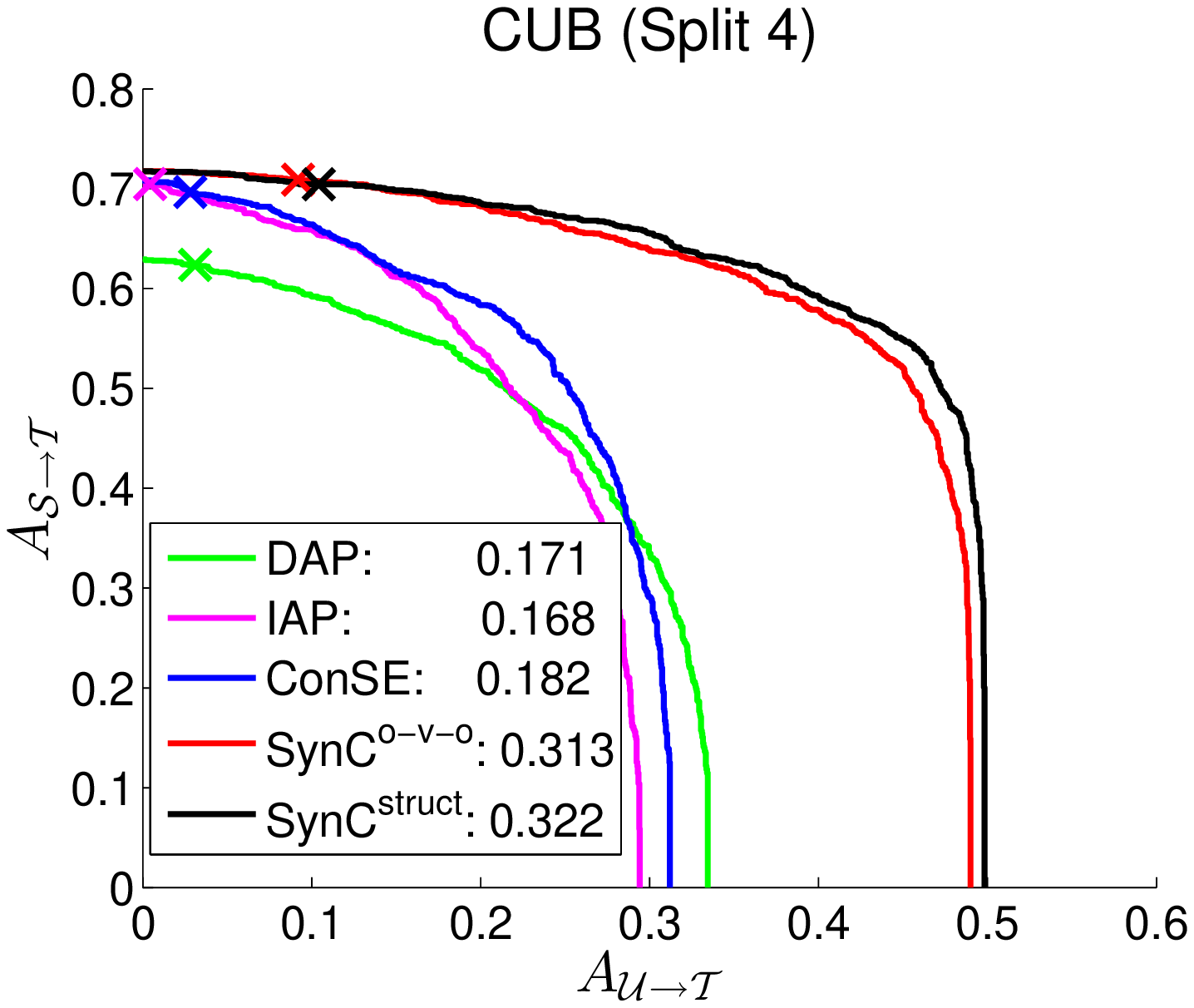}
\caption{\small Comparison of performance measured in AUSUC between different zero-shot learning approaches on the four splits of \textbf{CUB}.}
\label{fAUSUC_CUB}
\end{figure}

\begin{figure}[ht!]
\centering
\includegraphics[width=.46\textwidth]{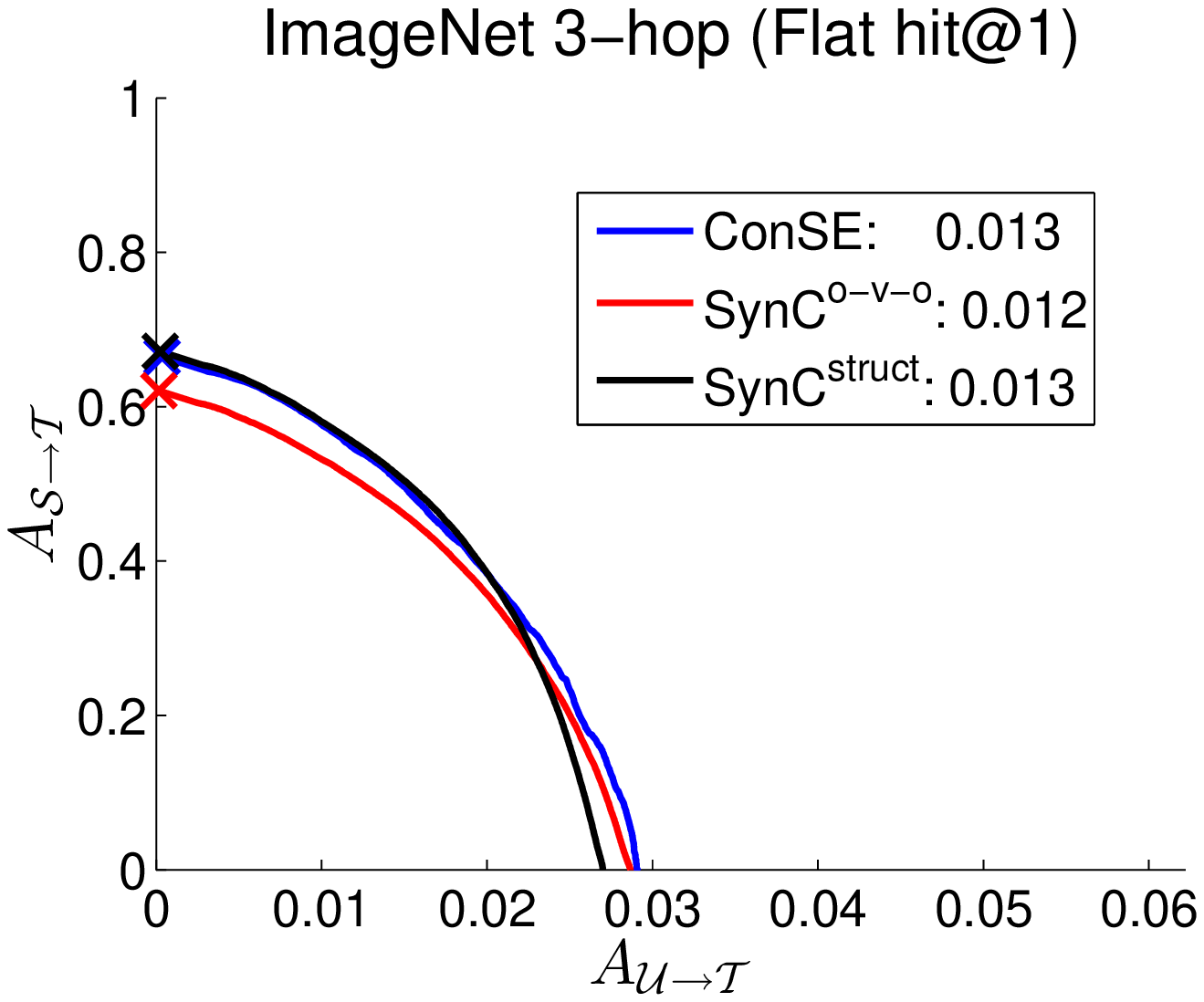}
\includegraphics[width=.46\textwidth]{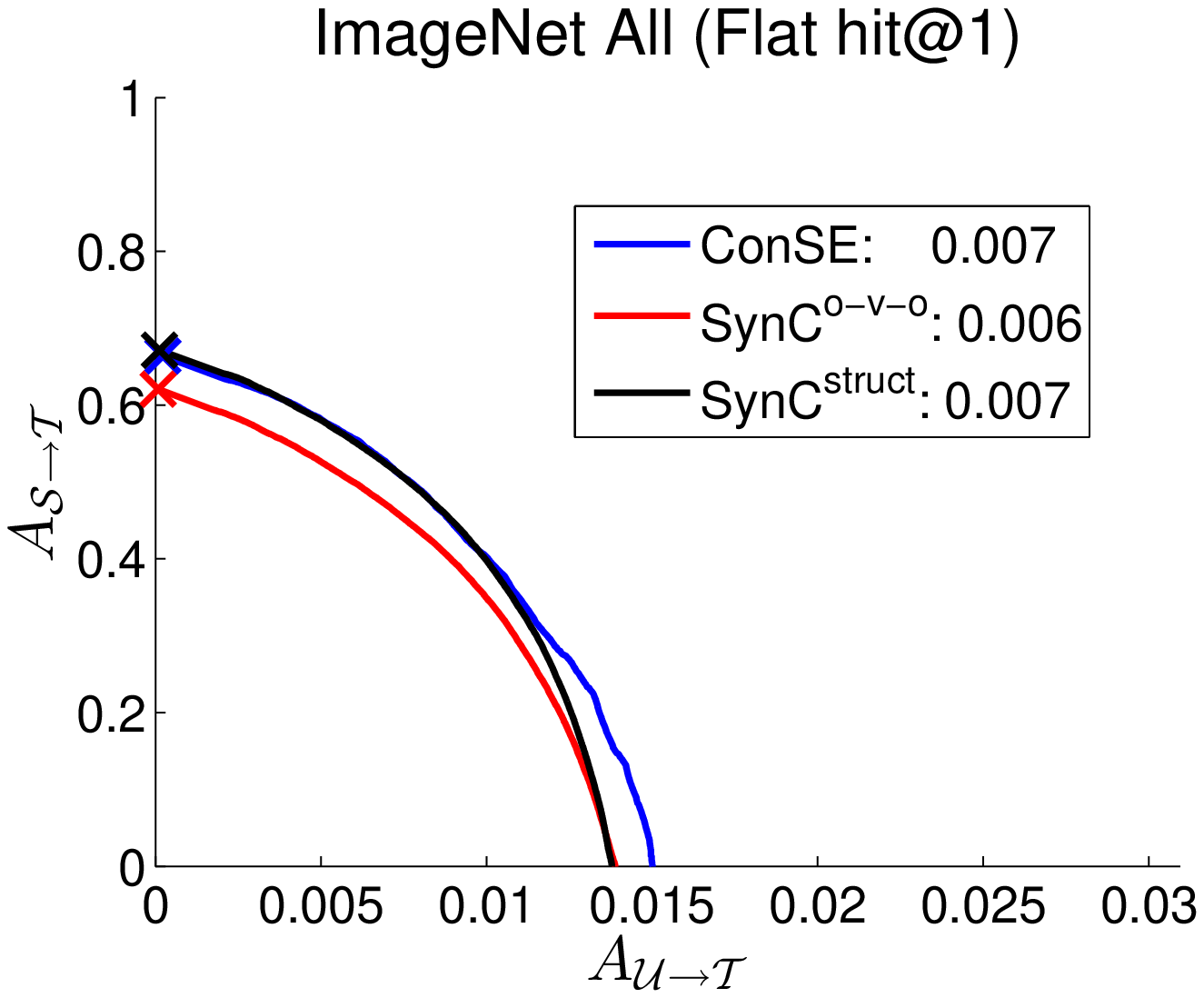}
\includegraphics[width=.46\textwidth]{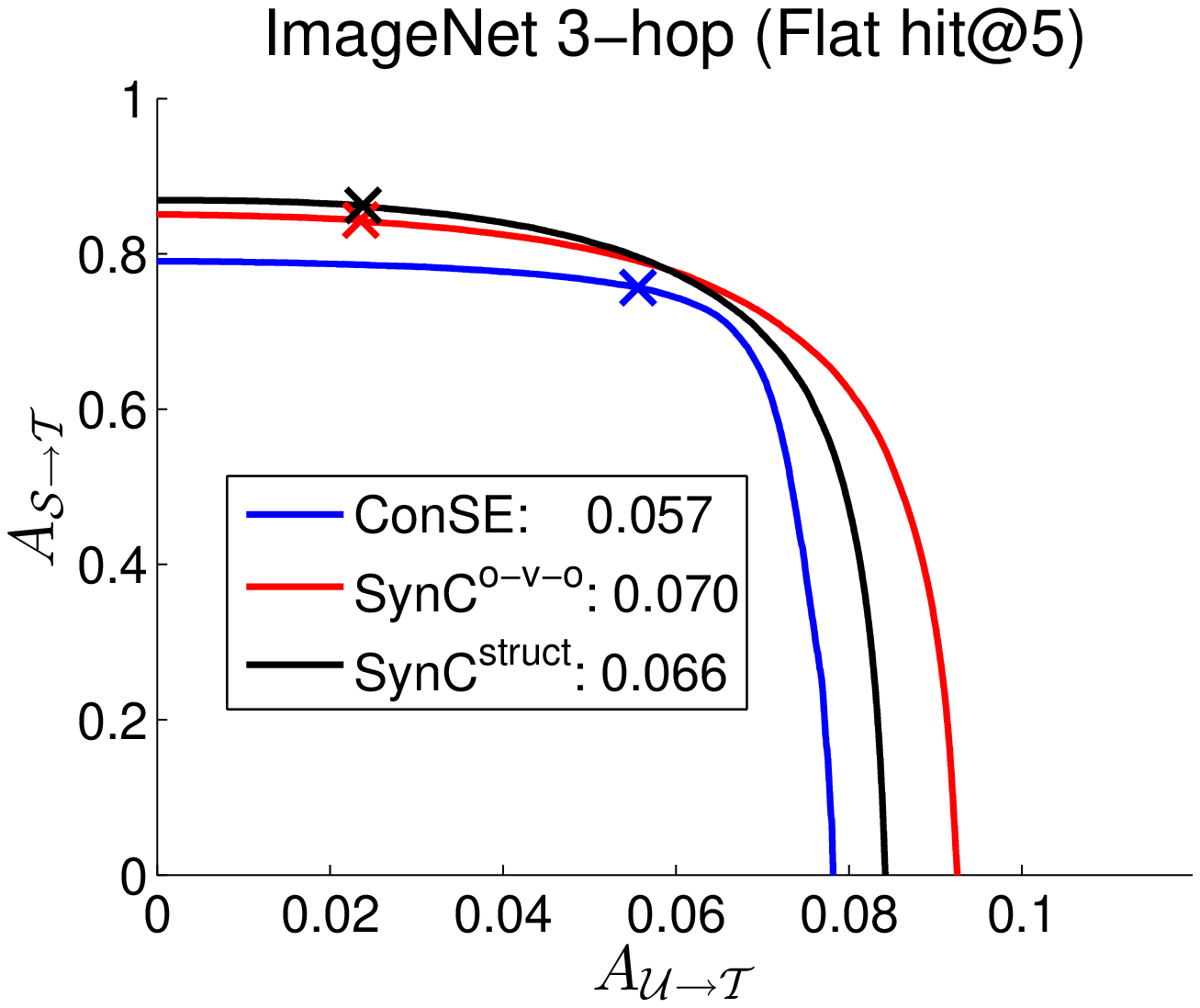}
\includegraphics[width=.46\textwidth]{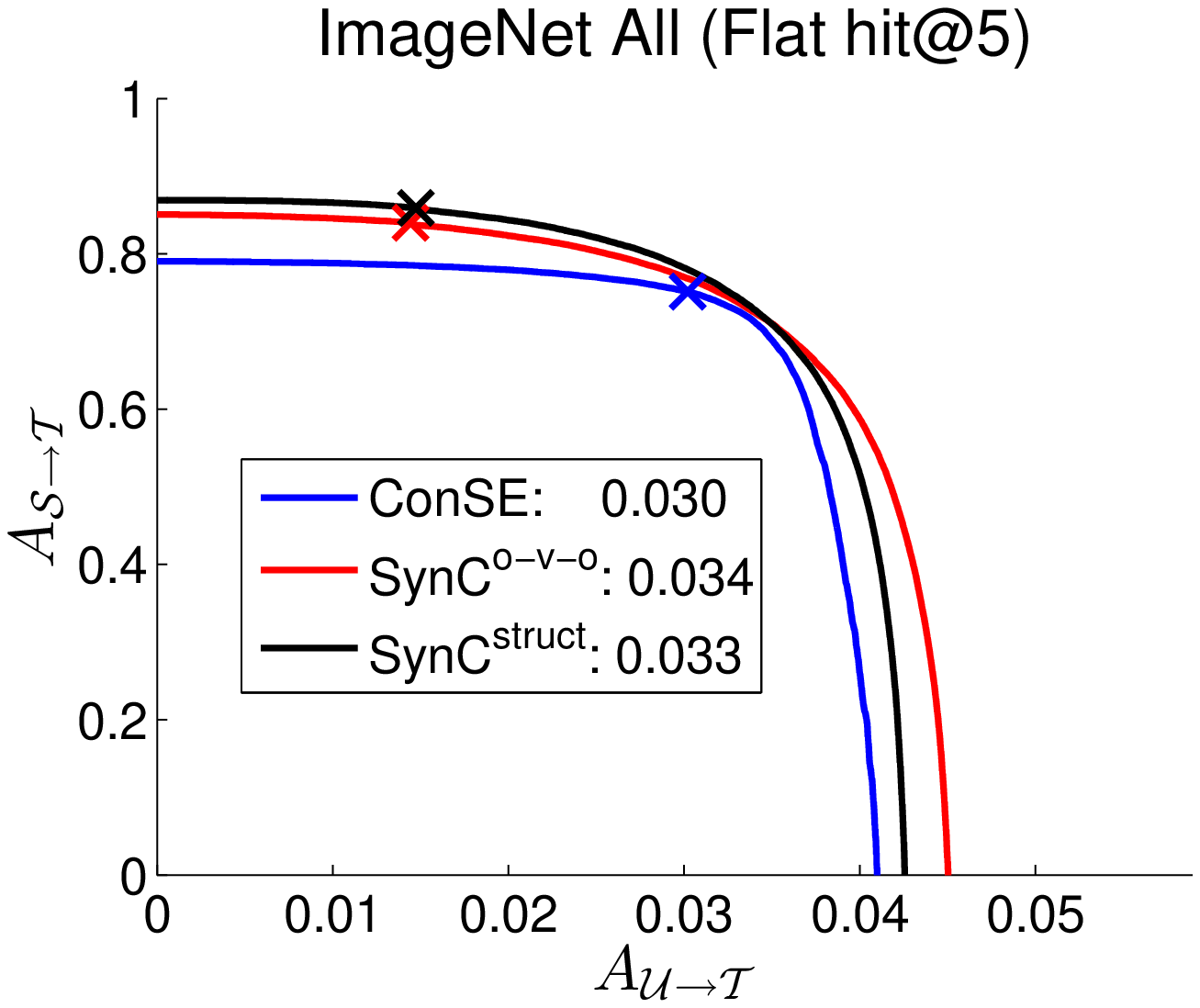}
\includegraphics[width=.46\textwidth]{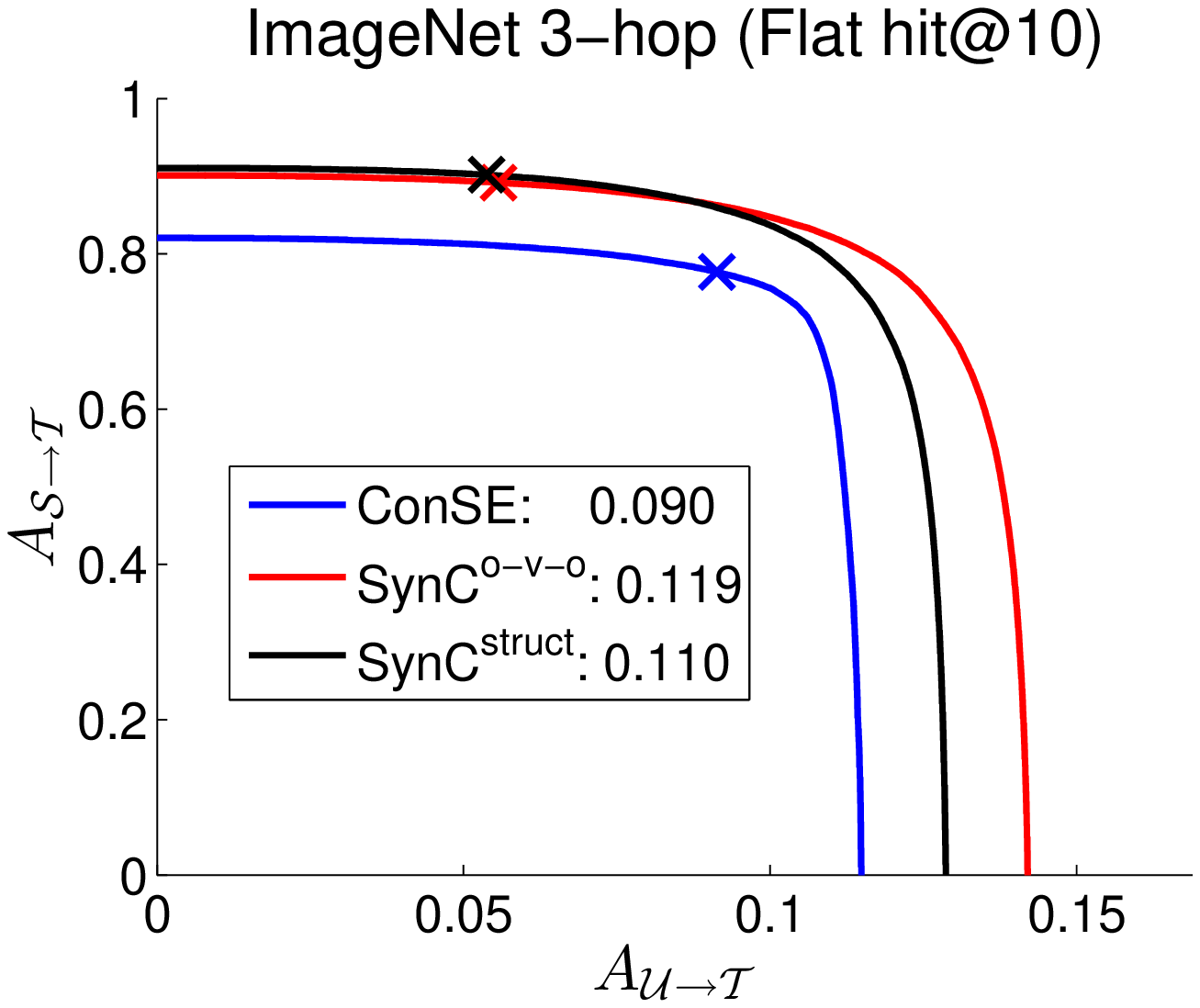}
\includegraphics[width=.46\textwidth]{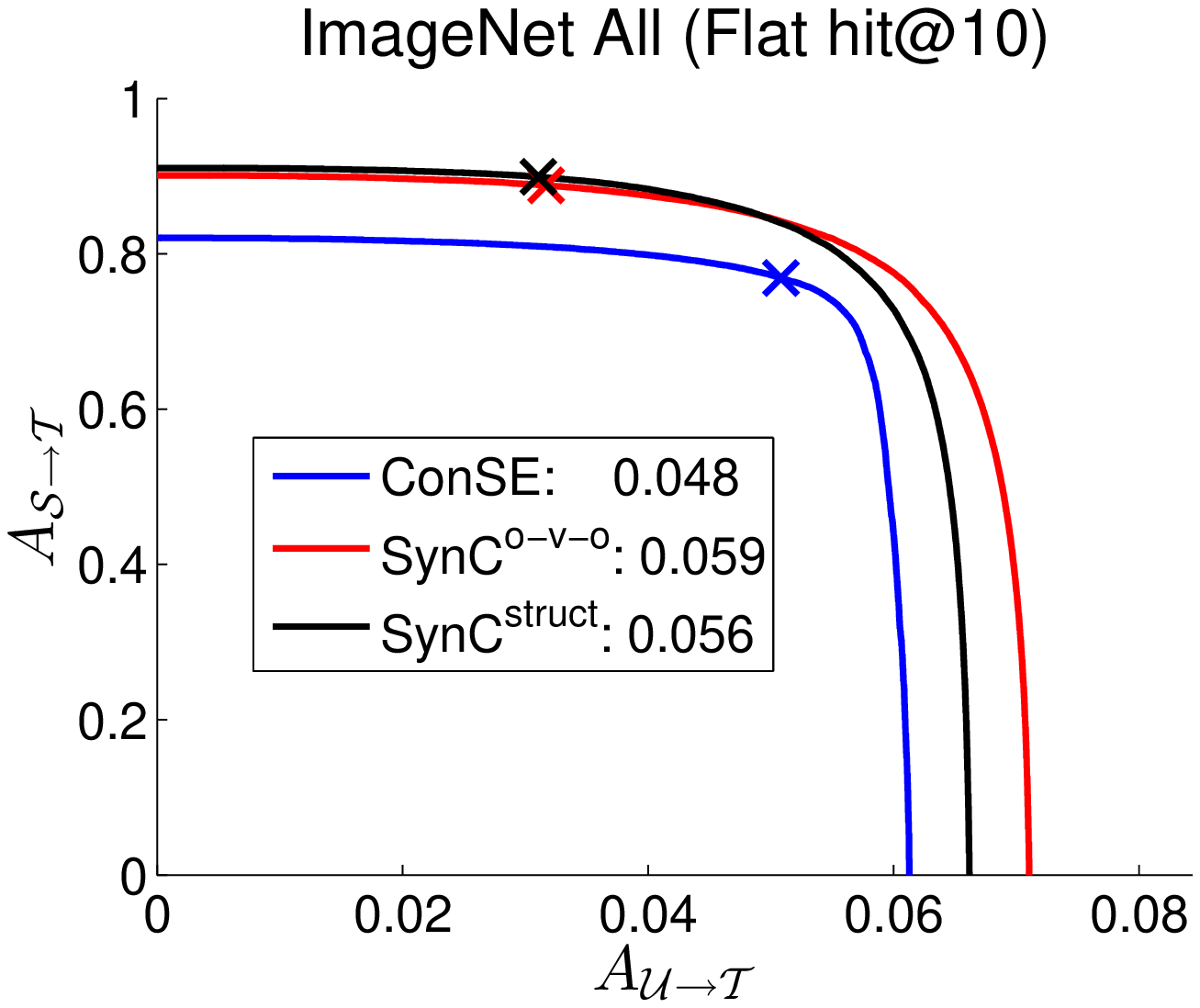}
\includegraphics[width=.46\textwidth]{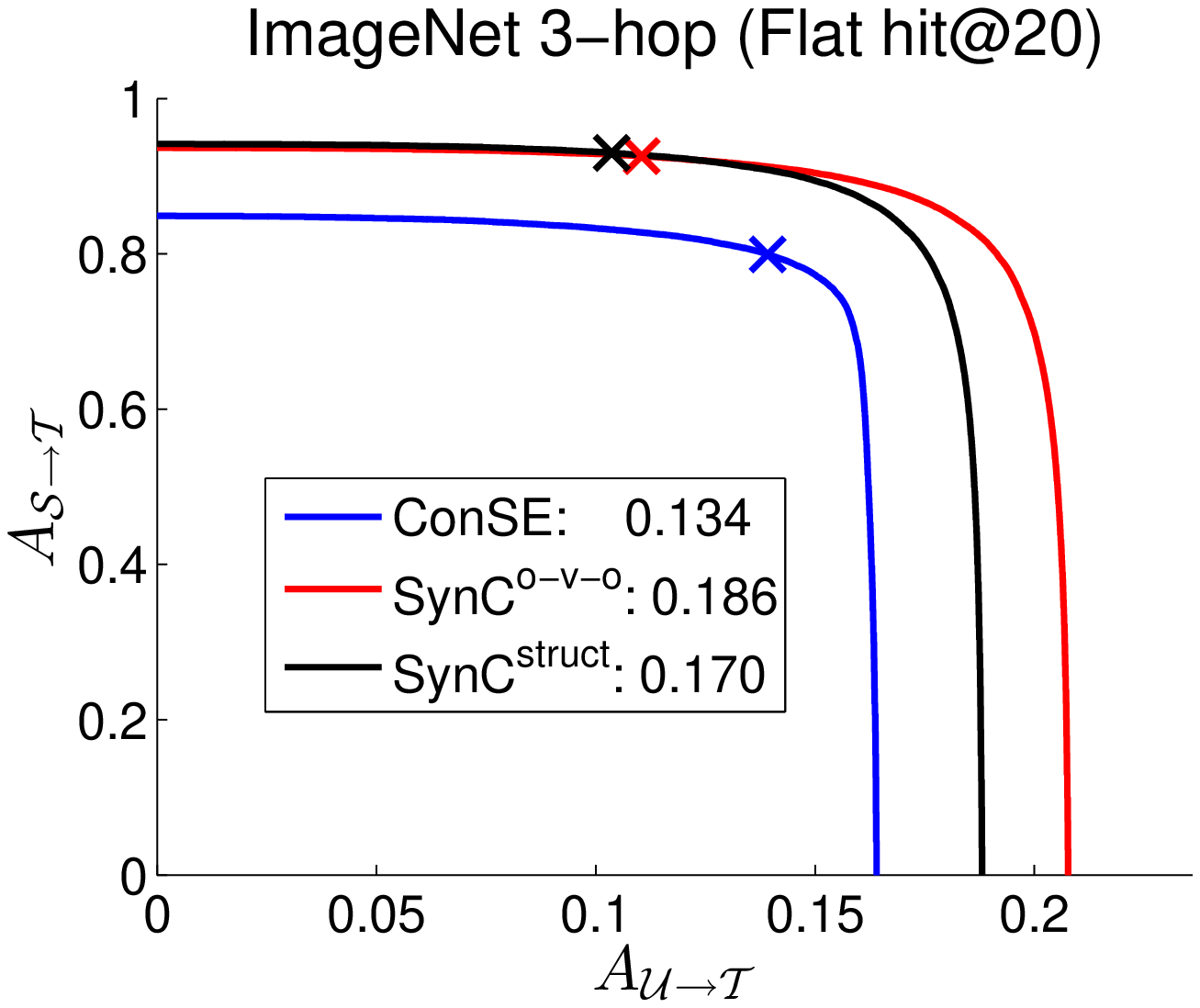}
\includegraphics[width=.46\textwidth]{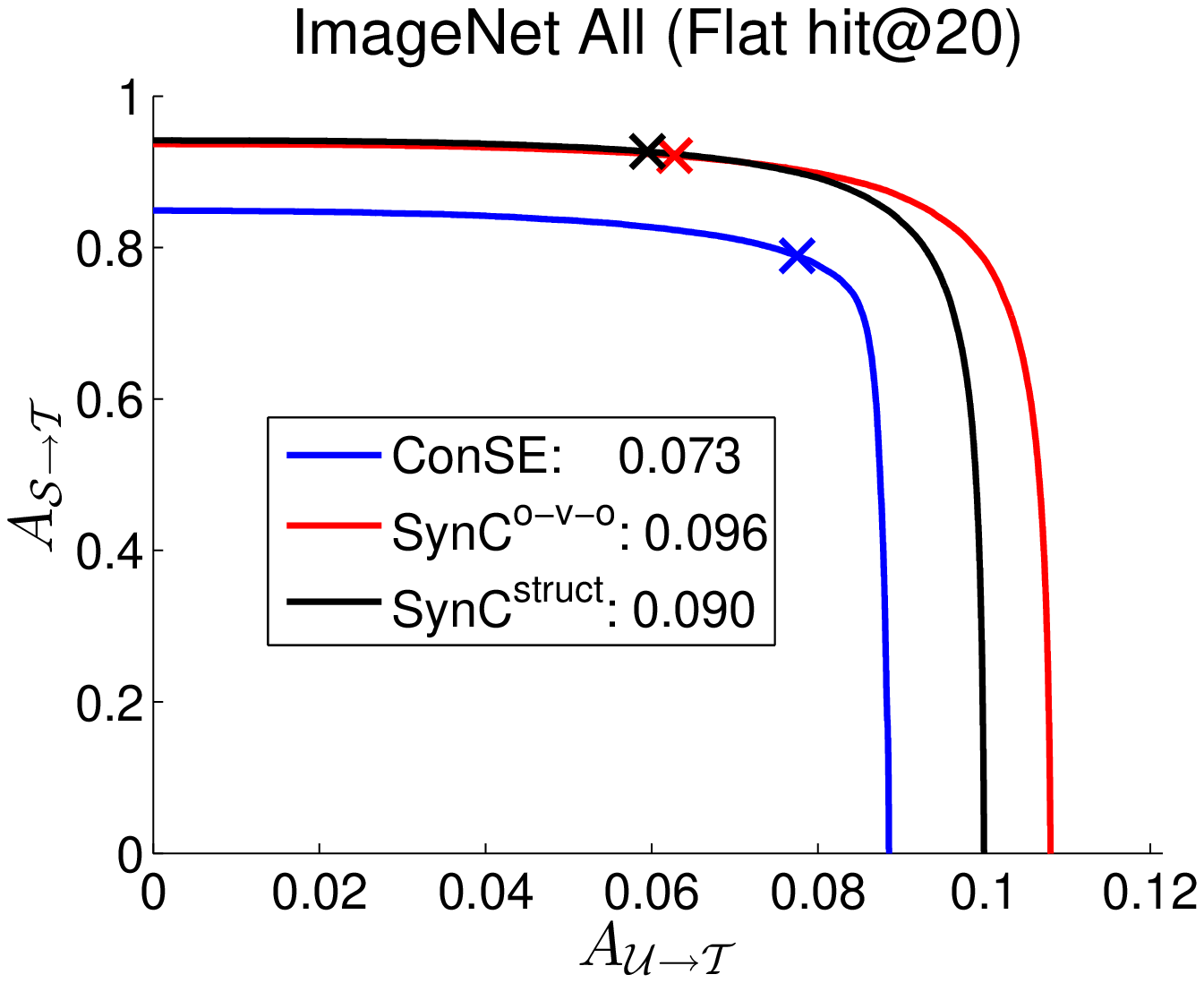}
\caption{\small Comparison of performance measured in AUSUC between different zero-shot learning approaches on \textbf{ImageNet-\emph{3hop}} (left) and \textbf{ImageNet-\emph{All}} (right).}
\label{fAUSUC_ImageNet_3hop_all}
\end{figure}

In particular, Fig.~\ref{fAUSUC_CUB} provides SUCs for all splits of \textbf{CUB} and Fig.~\ref{fAUSUC_ImageNet_3hop_all} provides SUCs for \textbf{ImageNet} \emph{3-hop} and \emph{All}. As before, we observe the superior performance of the method of SynC over other approaches in most cases. 

\subsection{Additional algorithm}
We examine an additional zero-shot learning algorithm ESZSL~\cite{Bernardino15} on the GZSL task. ESZSL learns a (linear) mapping from visual features to the semantic space (e.g., attributes or \textsc{word2vec}) with well-designed regularization terms. The mapped feature vector of a test instance is compared to semantic representations of classes (e.g., via dot products) for classification. Contrast to DAP/IAP~\cite{LampertNH14}, and similar to~\cite{AkataPHS13,AkataRWLS15}, ESZSL can easily incorporate continuous semantic representations and directly optimize the classification performance on seen classes. Table~\ref{tNov_ESZSL} shows the results on \textbf{AwA} and~\textbf{CUB}, with the hyper-parameters cross-validated to maximize AUSUC. Compared to other zero-shot learning algorithms (cf. Table 3 in the main text), ESZSL outperforms DAP/IAP/ConSE, but is still worse than both versions of SynC. Moreover, the proposed~\emph{calibrated stacking} clearly leads to better results than both novelty detection methods for ESZSL.

\begin{table}
\centering
{\small
\caption{Performance of ESZSL (measured in AUSUC) on \textbf{AwA} and \textbf{CUB}. Hyper-parameters are cross-validated to maximize AUSUC.}
\label{tNov_ESZSL}
\begin{tabular}{c|c|c|c||c|c|c} \hline
& \multicolumn{3}{|c||}{\textbf{AwA}} & \multicolumn{3}{|c}{\textbf{CUB}} \\ \cline{2-7}
Method  & \multicolumn{2}{|c|}{Novelty detection \cite{SocherGMN13}} & Calibrated  &  \multicolumn{2}{|c|}{Novelty detection \cite{SocherGMN13}} & Calibrated \\ \cline{2-3} \cline{5-6} 
 & Gaussian & LoOP & Stacking & Gaussian & LoOP & Stacking \\ 
\hline
ESZSL~\cite{Bernardino15} & 0.358 & 0.331 & 0.449 & 0.146 & 0.166 & 0.243\\
\hline
\end{tabular}
}
\end{table}

\subsection{Additional dataset}
We further experiment on another benchmark dataset, the \textbf{SUN attribute dataset (SUN)} \cite{PattersonH14}, which contains 14,340 images of 717 scene categories. We extract the 1,024 dimensional GoogLeNet features~\cite{SzegedyLJSRAEVR14} pre-trained on the ILSVRC 2012 1K training set~\cite{ILSVRC15}, and use the provided 102 continuous- and binary-valued attributes. Following~\cite{ChangpinyoCGS16,LampertNH14}, we split the dataset into 10 folds with disjoint classes (each with 71/72 classes), and in turn treat each fold as a test set of unseen classes. We then report the average results over 10 rounds. Similar to the setting on \textbf{AwA} and \textbf{CUB}, we hold out 20\% of the data points from seen classes and merge them with those of the unseen classes to form the test set. The results are summarized in Table~\ref{tNov_SUN}, where calibrated stacking outperforms Gaussian and LoOP for all ZSL algorithms. Besides, both versions of SynC still outperform the other ZSL algorithms.  

\begin{table}
\centering
{\small
\caption{Performance measured in AUSUC of several methods for Generalized Zero-Shot Learning on \textbf{SUN}. Hyper-parameters are cross-validated to maximize AUSUC.}
\label{tNov_SUN}
\begin{tabular}{c|c|c|c} \hline
& \multicolumn{3}{|c}{\textbf{SUN}} \\ \cline{2-4}
Method  & \multicolumn{2}{|c|}{Novelty detection \cite{SocherGMN13}} & Calibrated  \\ \cline{2-3} & Gaussian & LoOP & Stacking \\ 
\hline
DAP & 0.061 & 0.062 & 0.096 \\
IAP & 0.093 & 0.095 & 0.145 \\
ConSE & 0.120 & 0.122 & 0.200 \\
ESZSL & 0.017 & 0.018 & 0.026 \\
{SynC$^\textrm{o-vs-o}$} & 0.144 & 0.146 & 0.242 \\
{SynC$^\textrm{struct}$} & 0.151 & 0.153 & 0.260 \\
\hline
\end{tabular}
}
\end{table}

\section{Analysis on (generalized) zero-shot learning: details and additional results}
\label{ssAnalysis}

\subsection{Subsampling procedure for constructing \textbf{ImageNet-2K}}
In Section 6 of the main text, we construct \textbf{ImageNet-2K} by subsampling 1,000 unseen classes from the original 20,345 unseen classes of \textbf{ImageNet} to reduce computational cost. We pick 74 classes from \emph{2-hop}, 303 from ``pure" \emph{3-hop} (that is, the set of \emph{3-hop} classes that are not in the set of \emph{2-hop} classes), and 623 from the rest of the classes. These numbers are picked to maintain the proportions of the three types of unseen classes in the original \textbf{ImageNet} (see Evaluation Metrics in Section 5.1 of the main text for more details).
Each of these classes has between 1,050-1,550 examples.

\subsection{Additional results}

\begin{figure}
\centering
\includegraphics[width=.45\textwidth]{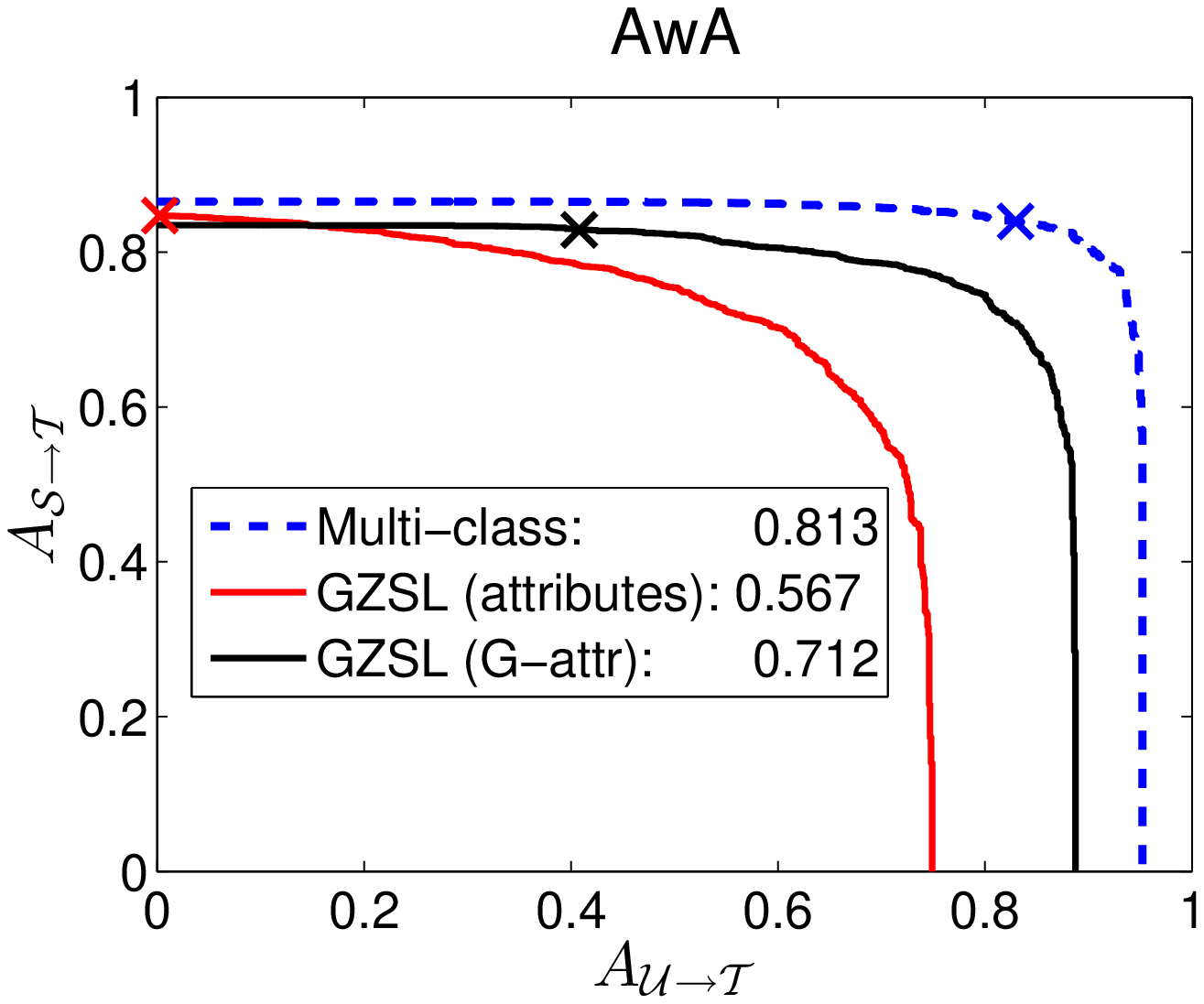}
\includegraphics[width=.45\textwidth]{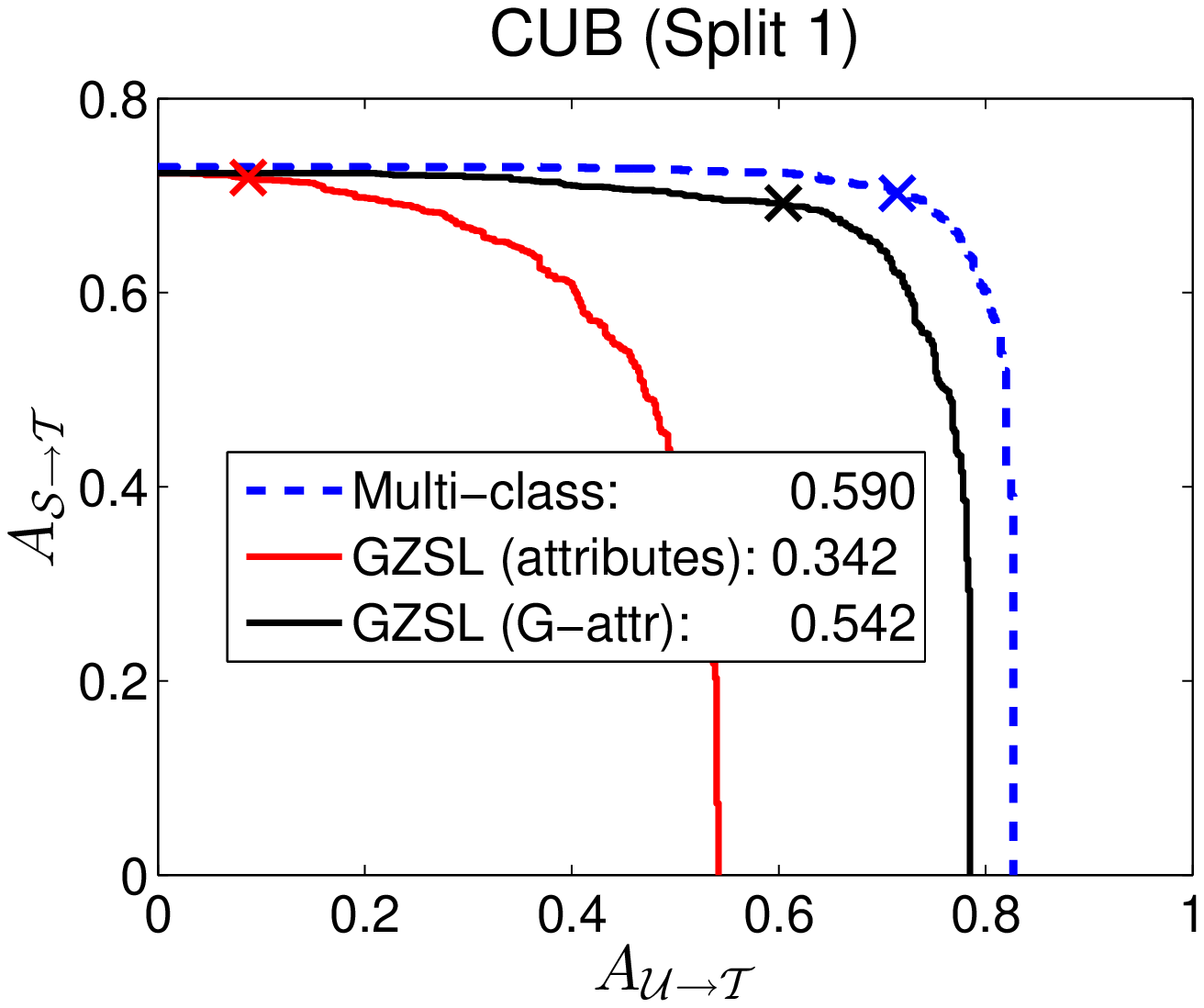}
\caption{\small Comparison between GZSL and multi-class classifiers trained with labeled data from both seen and unseen classes on the datasets \textbf{AwA} and \textbf{CUB}. GZSL uses visual attributes (in red color) or \textbf{G-attr} (in black color) as semantic embeddings.}
\label{fig:AWA_CUB_new}
\end{figure}

\begin{figure}
\centering
\includegraphics[width=.45\textwidth]{plots/I_1_new}
\includegraphics[width=.45\textwidth]{plots/I_5_new}
\includegraphics[width=.45\textwidth]{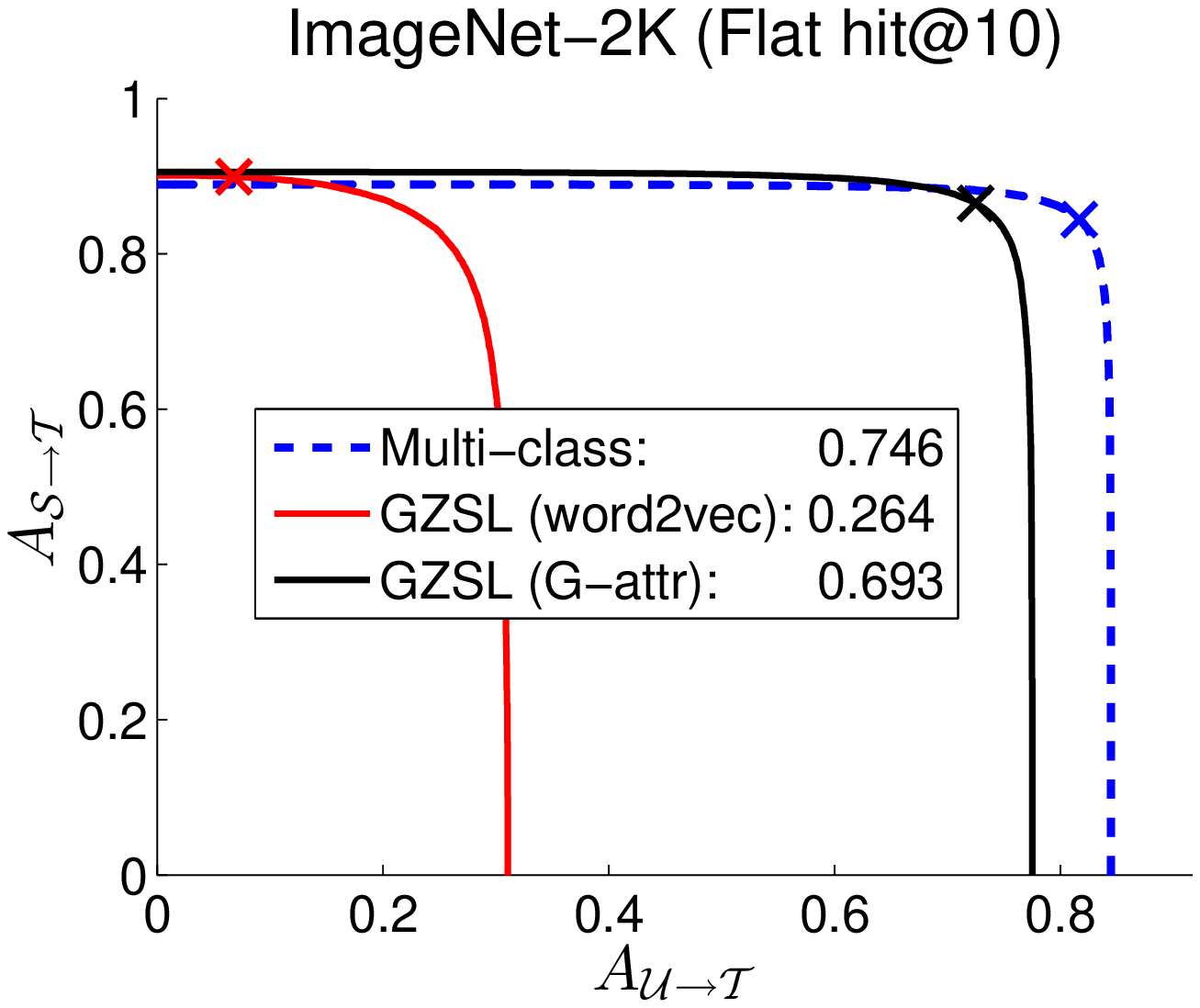}
\includegraphics[width=.45\textwidth]{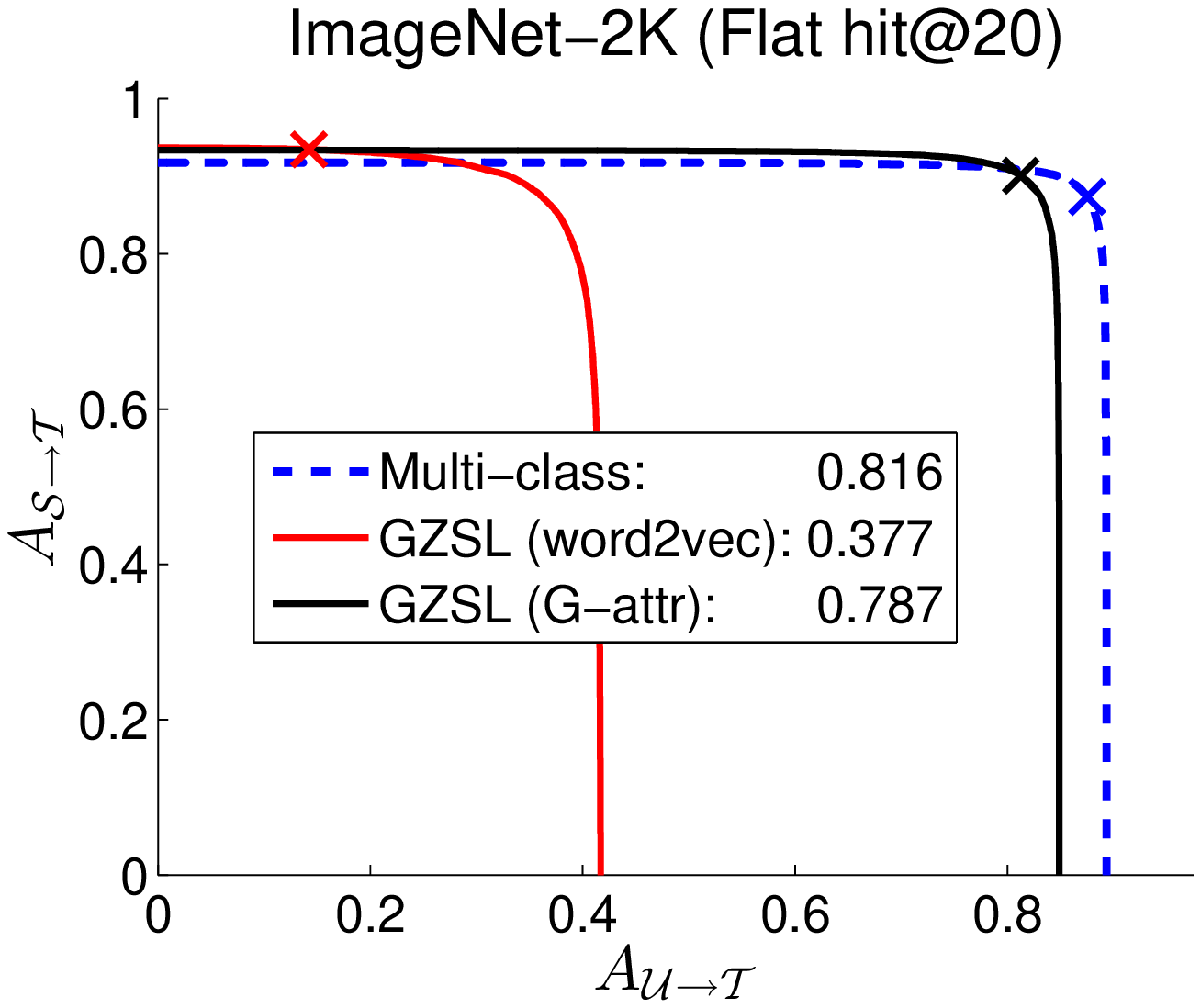}
\caption{\small Comparison between GZSL and multi-class classifiers trained with labeled data from both seen and unseen classes on the dataset \textbf{ImageNet-2K}. GZSL uses \textsc{word2vec} (in red color) or \textbf{G-attr} (in black color) as semantic embeddings.}
\label{fig:ImageNet_new_sup}
\end{figure}

We provide additional SUC plots on the performance of GZSL with idealized semantic embeddings \textbf{G-attr} in comparison to the performance of multi-class classification: Fig.~\ref{fig:AWA_CUB_new} for \textbf{AwA} and \textbf{CUB} and Fig.~\ref{fig:ImageNet_new_sup} for \textbf{ImageNet-2K}. As observed in Fig. 4 of Section 6 of the main text, the gaps between GZSL with visual attributes/\textsc{word2vec} and multi-class classifiers are reduced significantly.
The effect of \textbf{G-attr} is particularly immense on the \textbf{CUB} dataset, where GZSL almost matches the performance of multi-class classifiers without using labels from the unseen classes (0.54 vs. 0.59).

Finally, we provide additional results on the analysis of how much labeled data needed to improve GZSL's performance on \textbf{AwA} and \textbf{CUB}. In Table~\ref{detail_AWA_CUB}, we see the same trend as in Table 5 and 6 of the main text; GZSL with \textbf{G-attr} quickly approaches the performance of multi-class classifiers and large improvements from GZSL with visual attributes are observed --- even though these attributes are defined and annotated by human experts in this case. We also provide in Table~\ref{detail_ImageNet_suppl} and~\ref{detail_ImageNet_all_suppl} the full versions of Table 5 and 6 of the main text, with 2-shot and 5-shot results included.

\begin{table}
\centering
{\small
\caption{Comparison of performance measured in AUSUC between GZSL (using (human-defined) visual attributes and \textbf{G-attr}) and multi-class classification on \textbf{AwA} and \textbf{CUB}. Few-shot results are averaged over 1,000 rounds. GZSL with \textbf{G-attr} improves upon GZSL with visual attributes significantly. On \textbf{CUB}, the performance of GZSL with visual attributes almost matches that of multi-class classification.}
\label{detail_AWA_CUB}
\begin{tabular}{c|l|c||c} \hline
\multicolumn{2}{c|}{Method} & \textbf{AwA} &  \textbf{CUB} \\ \hline
& \multicolumn{1}{c|}{\textsc{Visual attributes}} & 0.57  & 0.34 \\ \cline{2-4}
& G-attr (1-shot) & 0.55\scriptsize{$\pm$0.04} & 0.26\scriptsize{$\pm$0.02} \\
& G-attr (2-shot) & 0.61\scriptsize{$\pm$0.03} & 0.34\scriptsize{$\pm$0.02} \\
GZSL & G-attr (5-shot) & 0.66\scriptsize{$\pm$0.02} & 0.44\scriptsize{$\pm$0.01} \\
& G-attr (10-shot) & 0.69\scriptsize{$\pm$0.02} & 0.49\scriptsize{$\pm$0.01} \\
& G-attr (100-shot) & 0.71\scriptsize{$\pm$0.003} & -$^\dagger$\\
& G-attr (all images) & 0.71 & 0.54\\  \hline
\multicolumn{2}{c|}{Multi-class classification} & 0.81 & 0.59 \\  \hline
\end{tabular}
}
\begin{flushleft}
$^\dagger$: We omit this setting as no class in CUB has more than 100 labeled images. 
\end{flushleft}
\end{table}

\begin{table}
\centering
{\small
\caption{Comparison of performance measured in AUSUC between GZSL (using \textsc{word2vec} and \textbf{G-attr}) and multi-class classification on \textbf{ImageNet-2K}. Few-shot results are averaged over 100 rounds. GZSL with \textbf{G-attr} improves upon GZSL with \textsc{word2vec} significantly and quickly approaches multi-class classification performance.}
\label{detail_ImageNet_suppl}
\begin{tabular}{c|l|c|c|c|c} \hline
\multicolumn{2}{c|}{Method} & \multicolumn{4}{|c}{Flat hit@K} \\ \cline{3-6}
\multicolumn{2}{c|}{}         & 1 & 5 & 10 & 20\\ \hline
& \multicolumn{1}{c|}{\textsc{word2vec}} & 0.04  & 0.17 & 0.27& 0.38\\ \cline{2-6}
& G-attr from 1 image  & 0.08\scriptsize{$\pm$0.003} & 0.25\scriptsize{$\pm$0.005} & 0.33\scriptsize{$\pm$0.005} & 0.42\scriptsize{$\pm$0.005}\\
& G-attr from 2 images & 0.12\scriptsize{$\pm$0.002} & 0.33\scriptsize{$\pm$0.004} & 0.44\scriptsize{$\pm$0.005} & 0.54\scriptsize{$\pm$0.005}\\
GZSL & G-attr from 5 images & 0.17\scriptsize{$\pm$0.002} & 0.44\scriptsize{$\pm$0.003} & 0.56\scriptsize{$\pm$0.003} & 0.66\scriptsize{$\pm$0.003}\\
& G-attr from 10 images & 0.20\scriptsize{$\pm$0.002} & 0.50\scriptsize{$\pm$0.002} & 0.62\scriptsize{$\pm$0.002} & 0.72\scriptsize{$\pm$0.002}\\
& G-attr from 100 images  & 0.25\scriptsize{$\pm$0.001} & 0.57\scriptsize{$\pm$0.001}& 0.69\scriptsize{$\pm$0.001} & 0.78\scriptsize{$\pm$0.001}\\
& G-attr from all images & 0.25 & 0.58& 0.69& 0.79\\  \hline
\multicolumn{2}{c|}{Multi-class classification} & 0.35 & 0.66 & 0.75& 0.82\\  \hline
\end{tabular}
}
\end{table}

\begin{table}
\centering
{\small
\caption{Comparison of performance measured in AUSUC between GZSL with \textsc{word2vec} and GZSL with \textbf{G-attr} on the full \textbf{ImageNet} with 21,000 unseen classes. Few-shot results are averaged over 20 rounds.}
\label{detail_ImageNet_all_suppl}
\begin{tabular}{l|c|c|c|c} \hline
\multicolumn{1}{c|}{Method} & \multicolumn{4}{|c}{Flat hit@K} \\ \cline{2-5}
        & 1 & 5 & 10 & 20\\ \hline
\multicolumn{1}{c|}{\textsc{word2vec}} & 0.006 & 0.034 & 0.059 & 0.096 \\ \hline
G-attr from 1 image & 0.018\scriptsize{$\pm$0.0002}& 0.071\scriptsize{$\pm$0.0007}& 0.106\scriptsize{$\pm$0.0009} &0.150\scriptsize{$\pm$0.0011}\\
G-attr from 2 images & 0.027\scriptsize{$\pm$0.0002}& 0.103\scriptsize{$\pm$0.0007}& 0.153\scriptsize{$\pm$0.0009} &0.212\scriptsize{$\pm$0.0012}\\
G-attr from 5 images & 0.041\scriptsize{$\pm$0.0002}& 0.152\scriptsize{$\pm$0.0006}& 0.221\scriptsize{$\pm$0.0007} &0.300\scriptsize{$\pm$0.0008}\\
G-attr from 10 images & 0.050\scriptsize{$\pm$0.0002}& 0.184\scriptsize{$\pm$0.0003}& 0.263\scriptsize{$\pm$0.0004} &0.352\scriptsize{$\pm$0.0005}\\
G-attr from 100 images & 0.065\scriptsize{$\pm$0.0001}& 0.230\scriptsize{$\pm$0.0002}& 0.322\scriptsize{$\pm$0.0002} &0.421\scriptsize{$\pm$0.0002}\\
G-attr from all images & 0.067 & 0.236 & 0.329 & 0.429 \\  \hline
\end{tabular}
}
\end{table}

\end{document}